\documentclass[lettersize,journal]{IEEEtran}
\usepackage{amsmath,amsfonts}
\usepackage{algorithmic}
\usepackage{algorithm}
\usepackage{array}
\usepackage[caption=false,font=normalsize,labelfont=sf,textfont=sf]{subfig}
\usepackage{textcomp}
\usepackage{stfloats}
\usepackage{url}
\usepackage{verbatim}
\usepackage{graphicx}
\usepackage{cite}
\usepackage{amsmath,amsfonts}
\usepackage{algorithmic}
\usepackage{algorithm}
\usepackage{array}
\usepackage{textcomp}
\usepackage{stfloats}
\usepackage{url}
\usepackage{verbatim}
\usepackage{graphicx}
\usepackage{array}
\usepackage{cite}
\usepackage{algorithm}
\usepackage{algorithmic}
\usepackage{multirow}
\usepackage{amsmath}
\usepackage{amssymb}
\usepackage{pifont}
\usepackage{amsmath}
\usepackage{newfloat}
\usepackage{graphicx}
\usepackage{subcaption} 
\usepackage{listings}
\usepackage{tabularx}
\usepackage{hyperref}
\usepackage{float}
\hypersetup{hidelinks,
	colorlinks=true,
	allcolors=black,
	pdfstartview=Fit,
	breaklinks=true}
\usepackage{amssymb}
\newcommand{\cmark}{\ding{51}}

\ifCLASSOPTIONcompsoc
  \usepackage[nocompress]{cite}
\else
  \usepackage{cite}
\fi

\usepackage{etoolbox}

\usepackage{amssymb}
\usepackage{mathtools}
\usepackage{textcomp}
\usepackage{gensymb}
\usepackage{siunitx}
\usepackage{algorithmic}
\usepackage{algorithm}
\usepackage{listings}
\usepackage{multirow}
\usepackage{boldline}
\usepackage{array}
\usepackage{graphicx}
\usepackage{lipsum}
\usepackage{setspace}
\usepackage{stfloats}
\newcolumntype{C}[1]{>{\centering\arraybackslash}p{#1}}

\usepackage{url}
\usepackage{ragged2e}
\usepackage{float}
\usepackage{caption}

\begin{document}

\title{OS-HGAdapter: Open Semantic Hypergraph Adapter for Large Language Models Assisted Entropy-Enhanced Image-Text Alignment}

\author{
Rongjun~Chen,
Chengsi~Yao,
Jinchang~Ren,~\IEEEmembership{Senior~Member,~IEEE,}
Xianxian~Zeng,
Peixian~Wang,
Jun~Yuan,~\IEEEmembership{Member,~IEEE,}
Jiawen~Li,~\IEEEmembership{Senior~Member,~IEEE,}
Huimin~Zhao,
Xu~Lu
\thanks{Corresponding author: Jinchang Ren, Xianxian Zeng}%
\thanks{Rongjun~Chen, Huimin~Zhao and Xu~lu is with School of Computer Science, Guangdong Polytechnic Normal University, 510665, China, and also with Guangdong Provincial Key Laboratory of Intellectual Property and Big Data (e-mail: chenrongjun@gpnu.edu.cn; zhaohuimin@gpnu.edu.cn; xulu@gpnu.edu.cn)}
\thanks{Chengsi~Yao, Peixian~Wang, Jun~Yuan, Jiawen~Li and Xianxian Zeng are with School of Computer Science, Guangdong Polytechnic Normal University, 510665, China (e-mail: chengsiyao@gpnu.edu.cn; wangpeixian@gpnu.edu.cn; yuanjun@gpnu.edu.cn; lijiawen@gpnu.edu.cn; zengxianxian@gpnu.edu.cn; ).}
\thanks{Jinchang Ren is with School of Computer Science, Guangdong
Polytechnic Normal University, 510665, China, and also with the
National Subsea Centre, Robert Gordon University, AB21 0BH Aberdeen,
U.K. (e-mail: jinchang.ren@ieee.org)}
}

\maketitle

\begin{figure*}[t]
  \centering
  \subfloat[Conventional Text-Image matching method]
  {\includegraphics[width=0.6\textwidth,trim= 10 269 130 10 ,clip]{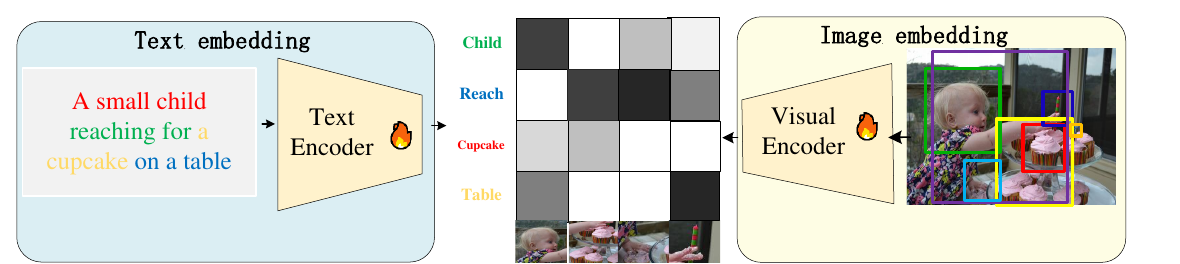}\label{fig:subfig1}}
  \subfloat[Human-brain Text-Image matching method]
  {\includegraphics[width=0.4\textwidth]{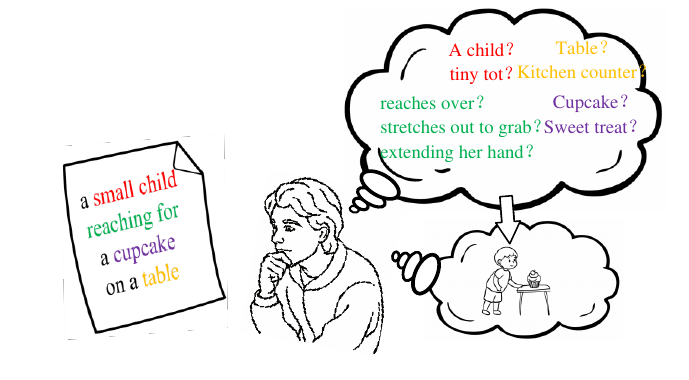}\label{fig:subfig2}}

  \subfloat[Our Text-Image matching method]
  {\includegraphics[width=1\textwidth,trim= 20 190 20 0 ,clip]{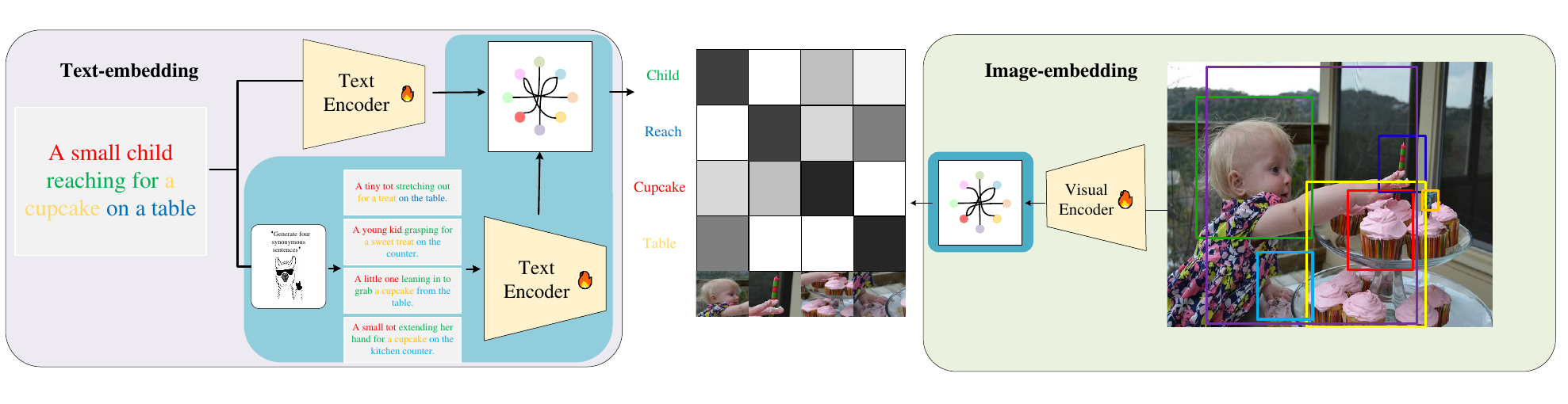}\label{fig:subfig3}}
  \caption{Comparison of different text-image alignment methods: (a) the regular way for matching the text and image, only following the original information of the modality. (b) Human's cross-modal ability is dominated by information and corrected by past cognition. The process of humans obtaining information from low-entropy modal information to the final result in our brain is entropy-increasing. The brain gradually fits and aligns information by increasing the entropy of memories\cite{quiroga2005invariant}. (c) Our image-text matching method imitates the pathway of entropy increase in the human brain through past memories. It simulates the entropy increase of human brain memory by prompting learning to guide the large model to output synonymous data. The text feature data after entropy increase is used to construct multi-link relationships using a hypergraph to calculate the endogenous relationship weights, and then sent to the similarity matrix for calculation. The blue highlights are the difference between (a) and (c)}
  \label{Figure1}
\end{figure*}

\begin{abstract}
Text-image alignment constitutes a foundational challenge in multimedia content understanding, where effective modeling of cross-modal semantic correspondences critically enhances retrieval system performance through joint embedding space optimization. Given the inherent difference in information entropy between texts and images, conventional approaches often show an imbalance in the mutual retrieval of these two modalities. To address this particular challenge, we propose to use the open semantic knowledge of Large Language Model (LLM) to fill for the entropy gap and reproduce the alignment ability of humans in these tasks. Our entropy-enhancing alignment is achieved through a two-step process: 1) a new prompt template that does not rely on explicit knowledge in the task domain is designed to use LLM to enhance the polysemy description of the text modality. By analogy, the information entropy of the text modality relative to the visual modality is increased; 2) A hypergraph adapter is used to construct multilateral connections between the text and image modalities, which can correct the positive and negative matching errors for synonymous semantics in the same fixed embedding space, whilst reducing the noise caused by open semantic entropy by mapping the reduced dimensions back to the original dimensions. Comprehensive evaluations on the Flickr30K \cite{young2014image} and MS-COCO \cite{chen2015microsoft} benchmarks validate the superiority of our Open Semantic Hypergraph Adapter (OS-HGAdapter), showcasing 16.8\% (text-to-image) and 40.1\% (image-to-text) cross-modal retrieval gains over existing methods while establishing new state-of-the-art performance in semantic alignment tasks.
\end{abstract}


\section{Introduction}
\IEEEPARstart{W}{ith} the development of the information age, various multimodal data are flooding our lives, and studying the characteristics and correlations between modalities becomes increasingly important. Text and images are two primary modalities of human cognition of the world, and the correlation between them has always been the focus of research. In recent years, image-text retrieval has become one kind of the hot spots of multi-modal research \cite{frome2013devise,lee2018stacked,li2022image}. Good alignment is directly related to correctly measuring the similarity between images and text, but the gap between modalities always has an unbridgeable barrier \cite{wu2022characterizing, huang2022modality}. Earlier work usually uses the global embedding \cite{frome2013devise} and key fragments \cite{chen2021learning,huang2017instance,li2019visual,Zhang2016Cross-modal} for alignment. In contrast, in the key fragment method, Stacked cross attention (SCAN) \cite{lee2018stacked}. These derived models \cite{diao2021similarity,liu2019focus,wu2019learning,zhang2022negative} employ local visual-textual associations to coordinate discriminative image features with textual elements, fusing contextual correspondences between region-word pairs for holistic cross-modal relevance assessment. These methods have always focused on optimizing the model itself, and paid little attention to the gap between modalities \cite{fan2023pmr}. However, the key challenge, namely bridging the gap between modalities and achieving cross-modal semantic correspondence, still needs to be solved. Obvious retrieval gaps are generally observed in the experimental results of the existing methods: the retrieval rankings from images to text will be higher than those of text—search rankings for images. After encoding, existing fragment alignment and global embedding methods fit image and text modalities according to probability or embedding size. The problem is that the encoded embedding space embeded by the text and visual encoder is anchored and cannot capture the inherent inconsistencies caused by multiplicity and sparse annotations. The issue is further exacerbated by the imbalance between different modalities \cite{fan2023pmr}. Determinism \cite{fu2023learning}, deviations fitting with visual modalities, amplify the sparse matching problem in the data set itself \cite{chun2022eccv,parekh2020crisscrossed}.

The heterogeneous information asymmetry observed in human multimodal perception systems, as depicted in Fig. \ref{Figure1},can be computationally modeled through mnemonic association mechanisms, where cross-modal binding energies derived from episodic memory traces compensate for perceptual discrepancy gradients. In order to introduce this cross-modal ability similar to associative ability into the model \cite{wei2022learning}, we introduced open semantic knowledge, which adds synonymous semantic expressions that have been understood by LLM based on the original corpus. By increasing the synonymous information entropy of text modalities, open semantics becomes the basis of computer association capabilities, and the embedding space position covered by a single classification is calibrated. We designed prompt statements differently from prompt learning way like Contrastive Language-Image Pre-Training (CLIP) \cite{radford2021learning} to prompt the LLM to output a synonymous vocab. However, the open semantic information output by the LLM and the original caption are used for ordinary clustering learning, which does not allow the model to correct the distorted embedding space. Even the noisy corpus generated by the LLM will interfere with enriching synonymous information. Inspired by \cite{lim2022hypergraph,wang2023multilateral}, hypergraphs are capable of more effectively capturing the intricate relationships present in multimodal data, we found that the hypergraphs and hypergraph neural networks are effective tools for aggregating multi-connected relationships. For this reason, we designed the hypergraph adapter to construct a multilateral semantic relationship beyond the pairwise feature relationship between the original corpus and the open semantic corpus. At the same time, the hypergraph adapter introduces open semantic entropy through dimensionality reduction fitting and only expands a single feature class in the embedding space and effectively reducing the open semantic noise generated by prompt learning without forwarding restrictions. To solve the difference in information complexity between modalities \cite{dong2022m5product}, we also introduced a hypergraph adapter in the visual modality to connect the multilateral semantic relationships with the features of the visual modality so that it will output with the text modality—the same contribution.
Our innovation lies in the following three folds:

(1) Use LLM to perform modal enhancement of low-entropy modes, improve the semantic richness of text modalities through the generation of synonymous sentences, and alleviate the sparse matching problem of data sets;

(2) Design a hypergraph adapter, construct multilateral connection semantics, and perform dimensionality reduction fitting to reduce open vocabulary noise and adjust the matching error in the embedding space;

(3) Extensive experiments conducted under standardized evaluation protocols on the Flickr30K \cite{young2014image} and MS-COCO \cite{chen2015microsoft} datasets validate our framework's efficacy, with the proposed OS-HGAdapter yielding 16.8\% (text-to-image) and 40.1\% (image-to-text) cross-modal retrieval performance enhancements compared to existing baselines.

\section{Related Work}
\subsection{Cross-modal image-text matching} Recent research strategies can be categorized into two main approaches: Embedding Space Matching and Scoring Mechanism Matching. Both aim to construct a shared multi-dimensional space to enable unified mapping and deep correlation exploration between images and texts. In Embedding Space Matching, research focuses on parallel processing of images and texts, utilizing deep learning networks to encode the raw data into high-dimensional vectors embedded in a common feature space. Semantic similarity is then assessed through the cosine similarity \cite{faghri2017vse++}. To enhance the expressiveness of embedding vectors, studies widely adopt Graph Convolutional Networks (GCNs) \cite{li2019visual,wang2020cross} and self-attention mechanisms \cite{wen2020learning,wu2019learning}, strengthening complex intra-sample semantic connections. The Visual Semantic Reasoning Network (VSRN) model \cite{li2019visual} introduces a semantic reasoning network to extract local image features and integrate them into a global representation. Researchers optimize the common subspace to reduce redundancy and noise, including seeking representative embedding spaces \cite{chun2021probabilistic,liu2021cross}, designing precise similarity measurement functions \cite{vendrov2015order,yang2022vision}, and leveraging vision-language pre-training techniques \cite{bao2021beit,wang2022image}. Cross-modal Hard Aligning Network (CHAN)\cite{pan2023fine} utilizes a hard alignment network that focuses on the most relevant alignment pairs. Hierarchical relation modeling framework (HREM) \cite{fu2023learning} constructs a hierarchical relational model to capture multi-level relationships. Multimodal Alignment-Guided Dynamic Token Pruning (MADTP) \cite{cao2024madtp} introduces MAG and DTP modules to reduce computational costs while maintaining performance. The Composition method for Object Relations and Attributes (CORA) \cite{pham2024composing} constructs hierarchical scene graphs to encode object-attribute configurations, employing an edge-connected topology where nodes represent visual entities and edges model relational dependencies. Meanwhile, the Linguistic-Aware Patch Slimming Framework (LAPS) \cite{fu2024linguistic} systematically detects semantically redundant image regions via linguistically-guided supervision and rectifies both semantic coherence and spatial alignment of these regions through adaptive feature recalibration.

\begin{figure*}[t]
\centering
\includegraphics[width=18cm]{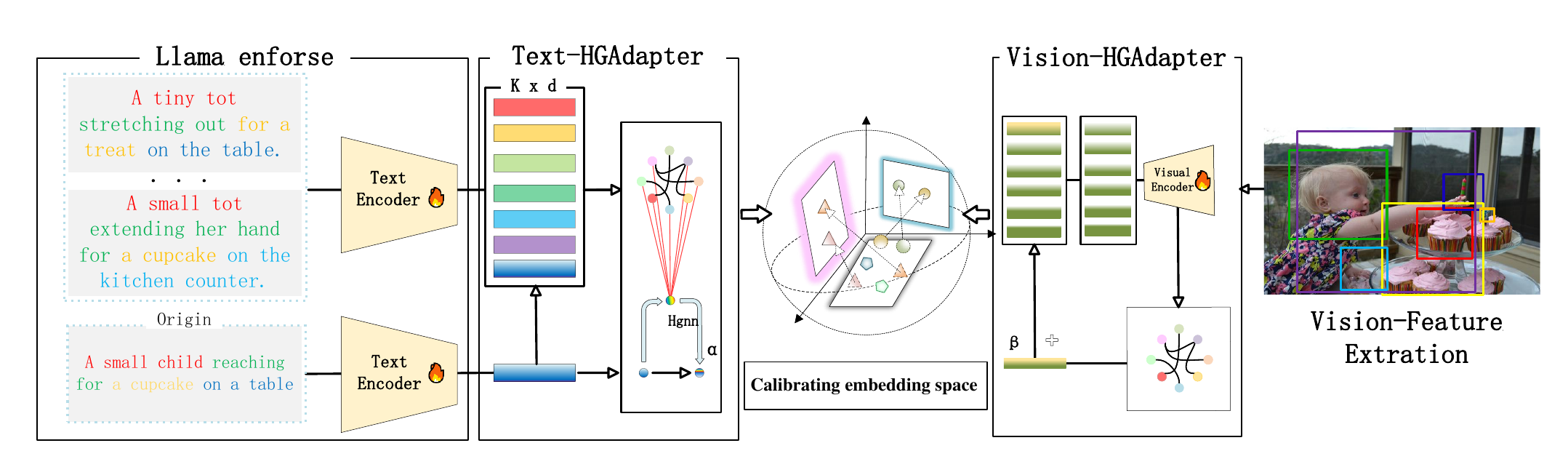} 
\caption{Overall structure of OS-HGAdapter. Which consists of in five main parts: The Llama enforce text encoder and Text-HGAdapter reanchor the text space, while visual feature extraction and Vision-HGAdapter capture the high-dimensional space of visual features to anchor the visual space. Finally, we integrate these into a Synonyms embedding space and perform calibration.}
\label{Figure2}
\end{figure*}

\subsection{Prompt Learning}Prompt-based Learning constitutes a meta-learning framework originating from NLP research, designed to optimize parameter-efficient transfer of LLM through task-aware template construction and contextual demonstration alignment, thereby enabling few-shot generalization across diverse downstream applications. In contrast to the traditional "Pre-training and Fine-tuning" strategy\cite{hu2021lora}, Prompt-based Learning leverages the construction and application of textual prompts to transform downstream tasks into forms more compatible with the pre-trained model, effectively reducing the domain shift between pre-training tasks and target downstream applications. Thus, it facilitates the smoother transition of knowledge accumulated during model pre-training to specific tasks. The earliest attempts at Prompt-based Learning involved the creation of templates using human prior knowledge \cite{petroni2019language}. At the same time, recent research has focused on applications in the discrete space of words \cite{jiang2020can,shin2020autoprompt} and the embedding space centered on sentence understanding. The Context Optimization (CoOP) \cite{zhou2022learning} and its extended versions have introduced Prompt-based Learning into open-world visual understanding, achieving significant performance improvements in few-shot visual scenarios. Meanwhile, Prompt learning with optimal transport (PLOT) \cite{chen2022plot} improves Prompt-based Learning by introducing the Optimal Transport distance to learn multiple local prompts, further enhancing fine-grained visual-language matching.

\section{The Proposed Methodology}
We propose OS-HGAdapter, a unified framework that integrates two synergistic components: (1) an LLM-driven synonymous sentence augmentation module for textual entropy mitigation, and (2) a dual-path hypergraph adapter for cross-modal feature refinement. As illustrated in Fig. \ref{Figure2}, this architecture employs modality-specific adapters—a textual adapter handling lexical generalization and a visual adapter stabilizing gradient alignment—to bridge information entropy gaps.

\subsection{Problem Formulation}

For each input image, we extract top-$K$ region-level features using an established visual encoder \cite{lee2018stacked}, specifically a Faster R-CNN backbone \cite{Ren2015Faster} pre-trained on Visual Genome \cite{Krishna2017Visual}. The architecture employs Bottom-Up/Top-Down attention (BUTD) \cite{Anderson2018Bottom-up} with multi-level aggregation for adaptive spatial contextualization. Features are projected into a $d$-dimensional shared embedding space via a dense layer, yielding a discriminative visual codebook:  

\begin{equation}V = \{ \nu_j \mid j \in [1, K],  \nu_j \in \mathbb{R}^d \},\end{equation}  
where $\nu_j$ denotes the $j$-th salient region embedding, $K$ is the total regions, and $d$ is the unified dimensionality.

For text encoding, we employ two approaches: a Bidirectional Gated Recurrent Unit (BiGRU) and a pre-trained BERT model \cite{Devlin2018Bert}. 
\begin{itemize}
    \item \textit{BiGRU-based encoder}: Each sentence $T$ is tokenized into words represented by pre-trained GloVe embeddings \cite{Pennington2014Glove}. These embeddings are processed through a BiGRU network to generate text queries $\mathcal{T} = \{ t_i \mid i \in [1, L],  t_i \in \mathbb{R}^d \}$, where $t_i$ encodes positional semantics of the $i$-th word and $L$ denotes sentence length. The final representation fuses forward and backward hidden states to capture bidirectional context.
    \item \textit{BERT-based encoder}: Token-level embeddings extracted from the final layer of a standard pre-trained BERT model leverage contextual embeddings for better semantic capture. They are projected to a $d$-dimensional latent space via a trainable linear layer.
\end{itemize}

Operationally, an information encoding process is employed, treating each word \( t_i \) as a query and the set of image features \( V = \{\nu_j\}_{j=1}^K \) as a visual codebook, with salient features serving as codewords. With reference to the cosine similarity \( s_{ij} = \frac{t_i^\top \hat{t_i}}{\|t_i\| \|\hat{t_i}\|} \), commonly adopted in cross-modal retrieval, the generalized representation on this codebook is defined as:
\begin{equation}\hat{t}_i = \sum_{j=1}^K \omega_{ij} \nu_j\label{eqVcodebook}\end{equation}
where \( \omega_{ij} \) denotes the weight coefficient for \( \nu_j \).

Standard probabilistic alignment often generates redundant correspondences owing to multiple candidate matches \cite{pan2023fine}. To mitigate this issue, CHAN \cite{pan2023fine} introduces Hard Assignment Coding, which exclusively selects the most relevant visual region for alignment. The alignment weight $\omega_{ij}$ is defined as:
\begin{equation}
\omega_{ij} = \begin{cases} 
1 & \text{if } j = \arg\max_k s_{ik}, \\
0 & \text{otherwise}.
\end{cases}
\end{equation}
Substituting Eq. (1) and Eq. (2), the text-visual similarity simplifies to:
\begin{equation}\label{eq3}
\begin{aligned}
s(t_{i}, \mathcal{V}) 
&= \frac{t_i^\top \hat{t_i}}{\|t_i\| \cdot \|\hat{t_i}\|} \\
&= \frac{t_i^\top \left( \sum_j \omega_{ij} \nu_j \right) }{\|t_i\| \cdot \left\| \sum_j \omega_{ij} \nu_j \right\|} \\
&= \frac{t_i^\top \nu_k}{\|t_i\| \cdot \|\nu_k\|} \text{(since } \omega_{ik} = 1 \text{ and } \omega_{ij} = 0 \text{ for } j \neq k\text{)} \\
&= s_{ik} = \max_{j=1,\dots,K} s_{ij},
\end{aligned}
\end{equation}
where $k = \arg\max_j s_{ij}$ denotes the index of the optimal visual codeword $\nu_k$.

Although this encoding framework retrieves relevant visual codewords for individual words, it fails to handle synonym alignment and semantic generalization in open-vocabulary embedding spaces, which manifests as lower text-to-image accuracy compared to image-to-text retrieval. The deterministic alignment—defined as a method that assigns each word to exactly one visual region based on maximum similarity—cannot recognize semantically equivalent expressions. For example, synonymous sentences \( T_1 \) and \( T_2 \) describing the same image \( \mathcal{V} \) may be misclassified as positive/negative pairs in frameworks like PCME++ \cite{chun2023improved}. This limitation is particularly evident in information-theoretic analysis\cite{chun2022eccv,parekh2020crisscrossed}, which attributes it to sparse cross-modal annotations constraining models such as CHAN \cite{pan2023fine}. While these frameworks optimize bidirectional triplet loss with online hard negative mining (VSE++ \cite{faghri2017vse++}), they cannot capture uncertainty from annotation multiplicity. Even with self-attention enhancements \cite{pan2023fine}, fixed-dimensional embeddings lack capacity to model complex synonym relationships, resulting in measurable semantic drift, characterized by positional deviations in the embedding space. Fundamentally, vision-language connections require probabilistic modeling beyond deterministic spaces.

\begin{figure}[t]
\centering
\small
\includegraphics[width=1\columnwidth]{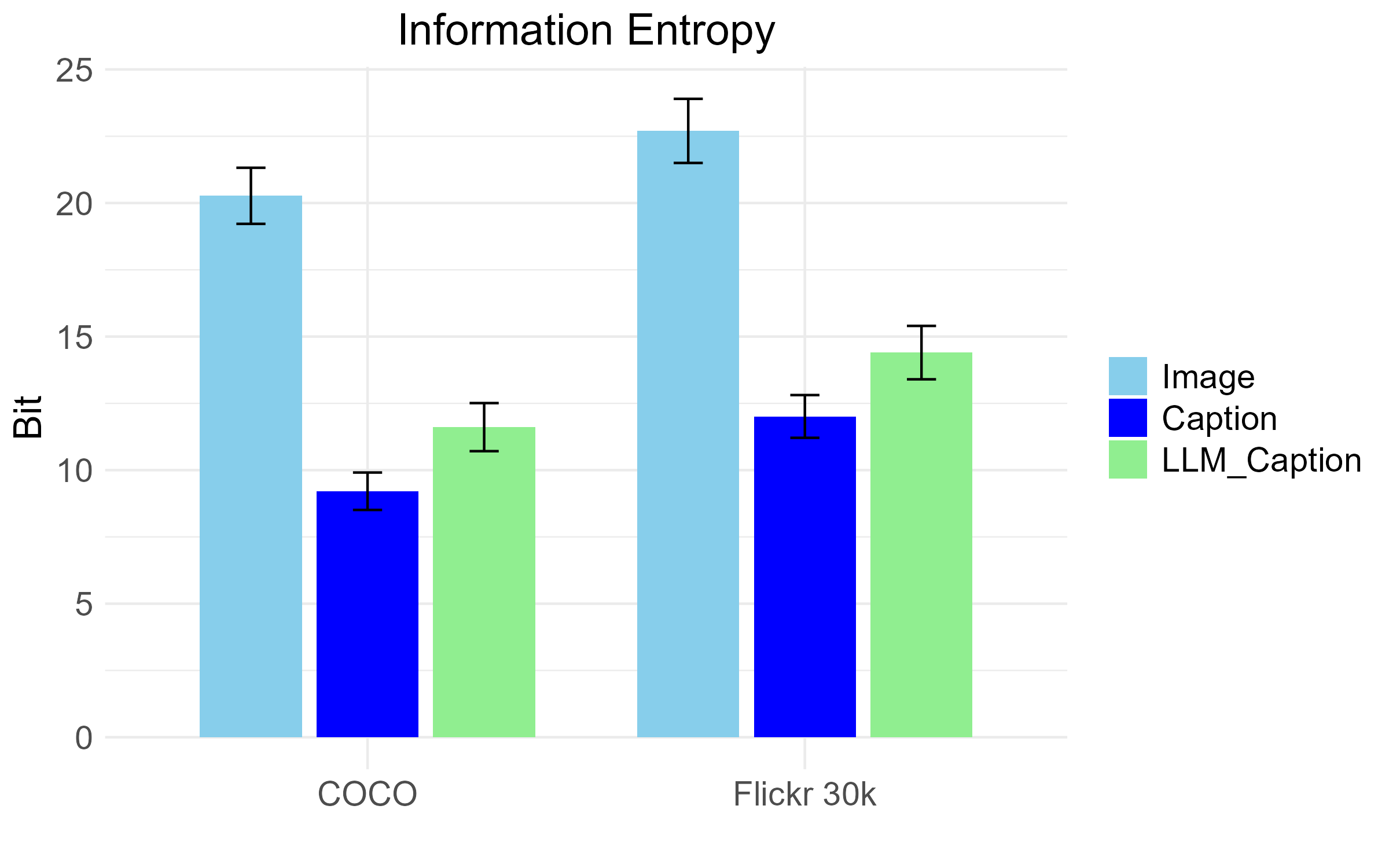} 
\caption{Comparing the information entropy values of images, text, and enhanced text on the two datasets.}
\label{Figure3}
\end{figure}

Multimodal tasks inherently exhibit significant information entropy disparities. Quantified via Shannon entropy over feature distributions (Fig. \ref{Figure3}), caption entropy averages $\ approx$9 bits versus $\ approx$20 bits for images across datasets. This divergence induces fundamental optimization asymmetry: low-entropy textual encodings compress semantic variability, while high-entropy visual features retain greater expressiveness. Crucially, unlike dynamically augmented text embeddings, the visual modality remains relatively static during joint training. This imbalance propagates substantial gradient misalignment during cross-modal optimization, accumulating irreversible encoding errors in visual-textual mapping paths that cannot be resolved through standard embedding adjustments. 

To address this fundamental challenge, OS-HGAdapter implements dual entropy mitigation strategies: textual entropy enhancement leverages LLM-generated synonymous sentences to expand lexical coverage, elevating textual entropy toward visual levels; concurrently, hypergraph correction employs a dynamic adapter path that iteratively adjusts the encoding space through structured feature interactions.

\subsection{LLM synonymous sentence reinforcement}

\begin{figure}[t]
\centering
\includegraphics[width=0.9\columnwidth]{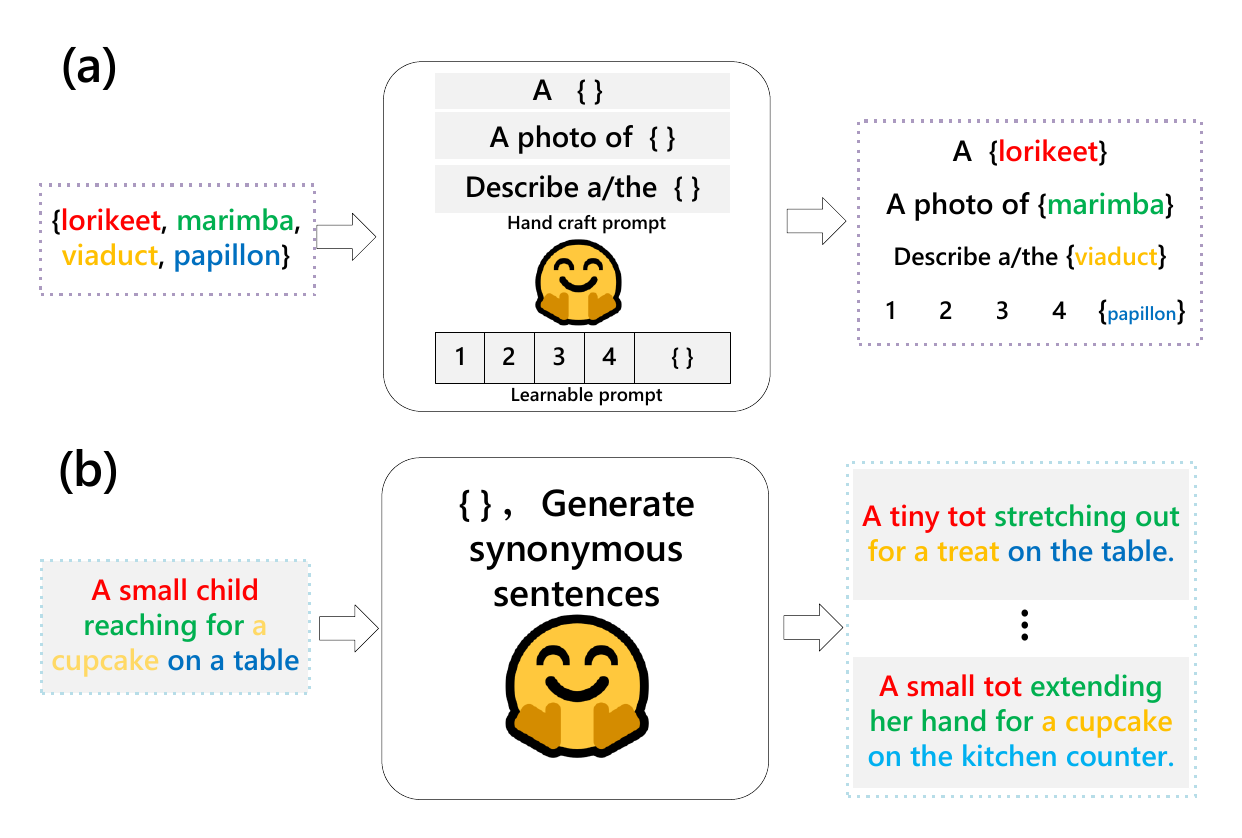} 
\caption{Results comparison from (a) template-based and adaptive prompt sentences, and (b) our prompt sentence}
\label{Figure4}
\end{figure}

We employ large language models (LLMs) to generate synonymous sentences by processing the dataset vocabulary \(T_{\text{Dataset}}\). Our prompt design, "\{\}, Generate synonymous sentences" (Fig. \ref{Figure4}(b)), preserves open generative capabilities, inducing synonymous semantics without constraining semantic diversity—unlike restrictive templates such as CLIP \cite{radford2021learning} or CUPL \cite{pratt2023does} (Fig. \ref{Figure4}(a)). This approach enables unconstrained vocabulary expansion, supporting cross-modal entropy regularization to bridge inter-modal entropy discrepancies and mitigate distributional distortions in the joint embedding manifold. Through LLM-based comprehension and regeneration, we enrich lexical coverage, adjust embedding space positions, and enhance phrase semantic understanding.  
To unify feature dimensions, we pad the LLM-generated vocabulary \(T_{\text{LLM}}\) with a separator token "[sep]" (represented as a zero tensor), ensuring \(l\) synonyms share a fixed dimension \(c\) (equal to the maximum word count after encoding). Given the original dataset vocabulary \(T_{\text{Dataset}}\) of dimensionality \(b\), we integrate \(c\) dimensions of synonymous information \(T_{\text{LLM}}(c)\), yielding a multi-dimensional representation \(F \in \mathbb{R}^{b+c}\):  
\begin{equation}  
F = \text{extend}(T_{\text{Dataset}}(b), T_{\text{LLM}}(c)).  
\end{equation}  
However, this augmentation introduces noise from inherent limitations in LLM comprehension. To mitigate such noise, we design the HG-Adapter, inspired by CLIP-Adapter \cite{gao2024clip}, which reduces embedding noise via weight control and relational connections.

\subsection{Hypergraph-Adapter}  
Learning synonymous information integration is critical for mitigating interference from unconstrained pre-trained language model outputs. To structurally integrate synonymous knowledge, we propose hypergraphs \( G = \{G_t\} \). These are constructed from the dataset corpus \( T_{Dataset} = \{\boldsymbol{t}_i \mid i \in [1, \dots, L], \boldsymbol{t}_i \in \mathbb{R}^b\} \) and the LLM-generated vocabulary \( T_{LLM} = \{\boldsymbol{t}_j \mid j \in [1, \dots, L], \boldsymbol{t}_j \in \mathbb{R}^c\} \). Original text features \( \boldsymbol{t}_{i=a} \) form multilateral semantic relations with other features \( \boldsymbol{t}_{i \neq a} \) and synonymous corpora \( T_{LLM} \), enabling distortion calibration in the deterministic embedding space toward an open embedding space.

Formally, each hypergraph \( H = (V, E, \mathbf{W}) \) comprises a vertex set \( V \), hyperedge set \( E \), and diagonal weight matrix \( \mathbf{W} \) encoding hyperedge-specific weights.  

Unlike conventional pairwise-edge graphs limited to binary connections, hypergraphs employ hyperedges to establish \( n \)-ary relations among nodes. Each hyperedge dynamically links multiple vertices with learned weights, modeling complex dependencies through overlapping node-edge interactions. This structure effectively captures high-dimensional relationships and heterogeneous semantics, which is critical for synonymous phrase modeling. The hypergraph topology is defined by an association matrix \( \mathbf{H} \in \mathbb{R}^{|V| \times |E|} \):  
\begin{equation}  
\mathbf{H}_{ij} =  
\begin{cases}  
1 & \text{if } v_i \in e_j \\  
0 & \text{otherwise}  
\end{cases}  
\end{equation}  
where the vertex degree \( d(v_i) = \sum_{j=1}^{|E|} \mathbf{H}_{ij} \) and hyperedge degree \( d(e_j) = \sum_{i=1}^{|V|} \mathbf{H}_{ij} \) quantify connectivity density.

To model semantically synonymous neighbors, the hypergraph adapter constructs hyperedges using the K-nearest neighbors (KNN) method \cite{gao2022hgnn+}, where each hyperedge \( e_i = \{v_i\} \cup N_{\mathrm{top-}k}(v_i) \) connects a vertex with its \( k \) most similar neighbors. This captures intrinsic feature relationships through the hyperedge set:
\begin{equation}
E_{\text{feature}}^{\mathrm{top-}k} = \{ N_{\mathrm{top-}k}(v) \mid v \in V \}
\end{equation}
The method efficiently scales to high-dimensional spaces by adaptively setting \( k \) values per search, enabling multifaceted connections between word features and visual codewords (defined in Eq. \ref{eqVcodebook}). Crucially, \( k \) modulates representation granularity and manifold topology of synonym embeddings.

We define the hyperparameter \(k\) for KNN-based hypergraph construction by leveraging feature activation characteristics. Specifically, \(k\) is set as the maximum dimension of text embeddings:  
\begin{equation}
k = \max(b, c)
\end{equation}
where \(b\) and \(c\) denote the dimensionality of \(\mathbf{T_{Dataset}}\) and \(\mathbf{T_{LLM}}\) embeddings, respectively. This configuration ensures broad coverage of the embedding topology. For each node \(v_i \in V\), a hyperedge is generated by combining \(v_i\) with its \(k\) nearest neighbors based on cosine similarity:  
\begin{equation}
e_i = \{v_i\} \cup \left\{ v_j \mid v_j \in \mathrm{top-}k \left( \frac{\mathbf{v}_i \cdot \mathbf{v}_j}{\|\mathbf{v}_i\| \|\mathbf{v}_j\|} \right) \text{ for } j \neq i \right\}
\end{equation}
where \(\mathrm{top-}n\) means sorting the similarity scores from high to low and selecting the nodes corresponding to the first \(n\) scores. \(b\) and \(c\) are the dimensions of \(\mathbf{T_{Dataset}}\) and \(\mathbf{T_{LLM}}\).

To model synonym relationships, we construct a hypergraph where each hyperedge connects synonymous word units. The weight matrix $\mathbf{W}$ integrates both original word units and synonym units through a diagonal block structure:
\begin{equation}
\mathbf{W} = \operatorname{diag}\left(\underbrace{w_o^1,\cdots,w_o^{n_o}}_{\text{original weights}}, \underbrace{w_s^1,\cdots,w_s^{n_s}}_{\text{synonym weights}}\right)
\end{equation}
where $w_o^i, w_s^j \in \mathbb{R}$ are learnable scalar weights associated with hyperedges, $n_o$ and $n_s$ denote the number of original word units and synonym units respectively. This diagonal formulation enables independent weight calibration for distinct semantic units during hypergraph propagation.

To model high-order semantic associations, we concatenate multiple hypergraph incidence matrices \cite{feng2019hypergraph}:
\begin{equation}
\mathbf{H} = \mathbf{H}_{\text{ori}} \parallel \mathbf{H}_{\text{sys}}^1 \parallel \cdots \parallel \mathbf{H}_{\text{sys}}^l
\end{equation}
where $\mathbf{H}_{\text{ori}} \in \{0,1\}^{|V| \times m_0}$ denotes the original hypergraph, $\mathbf{H}_{\text{sys}}^i \in \{0,1\}^{|V| \times m_i}$ represents the $i$-th synonym-based hypergraph, and $l$ is the number of synonym components. This concatenation preserves data independence while enriching potential semantic connections.

The hypergraph convolution is performed across \(K\) layers as:
\begin{equation}
\mathbf{F}^{(k+1)} = \sigma \left( 
\mathbf{D}_v^{-1/2} \mathbf{H} \mathbf{W} \mathbf{D}_e^{-1} \mathbf{H}^{\top} \mathbf{D}_v^{-1/2} \mathbf{F}^{(k)} \mathbf{\Theta}^{(k)}
\right)
\end{equation}
for \(k = 0, 1, \dots, K-1\). Here, \(\mathbf{\Theta}^{(k)} \in \mathbb{R}^{d_k \times d_{k+1}}\) is a trainable projection matrix, \(\sigma(\cdot)\) denotes an element-wise nonlinear activation function, and \(\mathbf{D}_v\) and \(\mathbf{D}_e\) are diagonal matrices representing vertex and hyperedge degrees, respectively.

\begin{figure}[t]
\centering
\includegraphics[width=0.8\columnwidth]{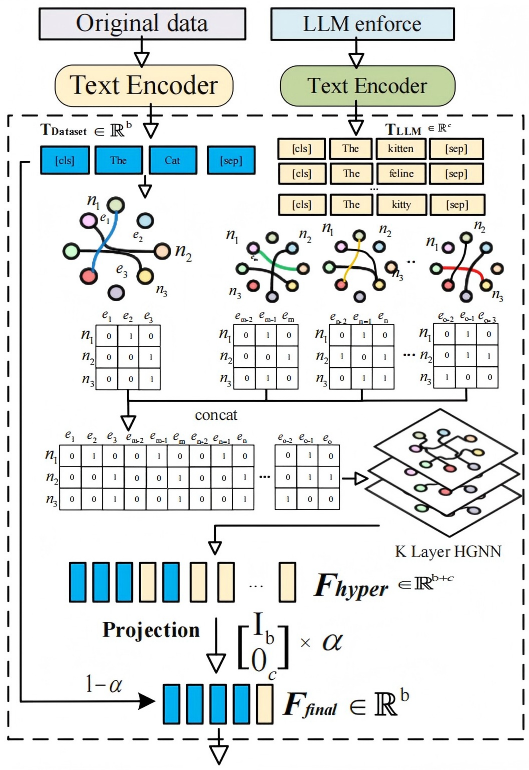} 
\caption{Hypergraph Adapter structure (dashed lines).}
\label{Figure5}
\end{figure}

The HGNN integrates features from the original image descriptions \( T_{Dataset} \in \mathbb{R}^b \) and LLM-generated features \( T_{LLM} \in \mathbb{R}^c \) into a joint representation \( F \in \mathbb{R}^{b+c} \). To preserve semantic coherence and mitigate noise in open-vocabulary data, we project \( F \) to the original dimensionality \( b \) via a linear operation:
\begin{equation}
\psi(F) = F \cdot A, \quad A = \begin{bmatrix} I_b & \mathbf{0}_{c} \end{bmatrix}^\top
\end{equation}
where \( I_b \) is the \( b \)-dimensional identity matrix. The final embedding, with \( \alpha \in [0,1] \) controlling the fusion ratio, combines this projection and residual connections to stabilize training:
\begin{equation}\label{eq:final_embedding}
F_{\text{final}} = (1 - \alpha) \cdot \psi(F) + \alpha \cdot T_{Dataset}
\end{equation}
This design ensures \( F_{\text{final}} \in \mathbb{R}^b \). The model minimizes the metric divergence \( \mathcal{L}_{\text{div}} \) between the deterministic feature space and open-vocabulary manifold during training.

The fusion ratio \(\alpha\) is determined by the normalized mutual information (NMI) between the original features \(T_{Dataset}\) and hypergraph-enhanced features \(\psi(F)\). This quantifies the shared information while accounting for dimensionality differences:
\begin{equation}
I(T_{Dataset}; \psi(F)) = \mathbb{E}_{p(x,y)} \left[ \log \frac{p(x,y)}{p(x)p(y)} \right]
\end{equation}
where \(x\) and \(y\) denote the continuous feature values. The NMI-based fusion ratio is:
\begin{equation}
\alpha = \frac{2 \cdot I(T_{Dataset}; \psi(F))}{h(T_{Dataset}) + h(\psi(F))}
\end{equation}
where \(h(X) = -\int p(x) \log p(x)  dx\) is the differential entropy. This formulation ensures \(\alpha \in [0,1]\) reflects the proportion of recoverable low-dimensional information.

The normalized mutual information ratio \(\alpha\) regulates hypergraph integration, where higher \(\alpha\) values amplify hypergraph influence (validated in Fig.~\ref{Figure7}). This adapter preserves information entropy while enhancing representation capacity.

Unlike the dynamically updated text embeddings, the visual modality \(V \in \mathbb{R}^d\) remains static. This asymmetry induces gradient deviation during cross-modal alignment:
\begin{equation}
\nabla_{\text{dev}} = \left\| \frac{\partial \mathcal{L}}{\partial V} - \frac{\partial \mathcal{L}}{\partial T} \right\|_2\label{dev}
\end{equation}
where \(\mathcal{L}\) denotes the alignment loss. To mitigate this deviation, we introduce a visual hypergraph adapter with residual connections, analogous to the text-side fusion controlled by \(\alpha\):
\begin{align}
V^{(t+1)} &= \beta \cdot \sigma\left( \mathbf{D}_v^{-1/2} \mathbf{H} \mathbf{W} \mathbf{D}_e^{-1} \mathbf{H}^{\top} \mathbf{D}_v^{-1/2} V^{(t)} \mathbf{\Theta}^{(t)} \right) \nonumber \\
          &\quad + (1 - \beta) \cdot V^{(t)}
\end{align}
Here \(\beta \in [0,1]\) is a fusion ratio analogous to \(\alpha\), which stabilizes cross-modal feature interactions through iterative refinement.

\begin{table*}[t]
\centering
\small
\caption{Test results of different models using different visual and language encoders on the coco 5k dataset and coco 5-fold 1k test set. Use \textcolor{red}{red} to highlight the best RSUM. }
 \resizebox{\textwidth}{!}{
\begin{tabular}{@{}llllllll|lllllll@{}}\hline
\multirow{3}{*}{Methods}      & \multicolumn{7}{c}{COCO 5-fold 1K Test}           & \multicolumn{7}{c}{COCO 5K Test}                  \\     
 &
  \multicolumn{3}{l}{I-\textgreater{}T} &
  \multicolumn{3}{l}{T-\textgreater{}I} &
  \multirow{2}{*}{RSUM} &
  \multicolumn{3}{l}{I-\textgreater{}T} &
  \multicolumn{3}{l}{T-\textgreater{}I} &
  \multirow{2}{*}{RSUM} \\
                             & R@1  & R@5  & R@10  & R@1  & R@5  & R@10  &       & R@1  & R@5  & R@10  & R@1  & R@5  & R@10  &       \\\hline
\textbf{Region+BiGRU}                &      &      &       &      &      &       &       &      &      &       &      &      &       &       \\\hline
SCAN\cite{lee2018stacked}             & 72.7 & 94.8 & 98.4  & 58.8 & 88.4 & 94.8  & 507.9 & 50.4 & 82.2 & 90.0  & 38.6 & 69.3 & 80.4  & 410.9 \\
VSE$\infty$ \cite{chen2021learning}            & 76.5 & 95.3 & 98.5  & 62.9 & 90.6 & 95.8  & 519.6 & 56.6 & 83.6 & 91.4  & 39.3 & 69.9 & 81.1  & 421.9 \\
CHAN\cite{pan2023fine}             & 79.7 & 96.7 & 98.7  & 63.8 & 90.4 & 95.8  & 525.0 & 60.2 & 85.9 & 92.4  & 41.7 & 71.5 & 81.7  & 433.4 \\
PCME++\cite{chun2023improved} & 81.9    & 97.1    & 98.9    & 69.4 & 92.8 & 97.1  & 537.4 &   62.7   &   86.6   &  93.2     & 47.9     &   76.6   &  85.7     &    452.7   \\
CGMN \cite{cheng2022cross}          & 76.8 & 95.4 & 98.3  & 63.8 & 90.7 & 95.7  & 520.7 & 58.9 & 85.2 & 92.0  & 41.4 & 71.6 & 82.6  & 431.7 \\
HREM\cite{fu2023learning}             & 81.2 & 96.5 & 98.9  & 63.7 & 90.7 & 96.0  & 527.1 & 60.6 & 86.4 & 92.5  & 41.3 & 71.9 & 82.4  & 435.1 \\
CORA\cite{pham2024composing}             & 81.7 & 96.7 & 99.0  & 66.0 & 92.0 & 96.7  & 532.1 & 63.0 & 86.8 & 92.7  & 44.2 & 73.9 & 84.0  & 444.6 \\
TVRN\cite{Pang2024mutually}        & 79.7 & 96.0 & 98.6   & 64.2 & 90.7 & 96.1  & 525.3  &   59.2   &   84.6   &  91.6     & 42.5     &   71.8   &  82.1     &    431.8 \\  
OS-HGAdapter(ours)            & 86.7 & 98.9 & 100.0  & 83.9 & 99.6 & 99.9 & \textcolor{red}{569.0} & 87.7 & 99.1 & 99.8  & 79.2 & 98.7 & 99.7  & \textcolor{red}{564.2} \\\hline
\textbf{Region+BERT}                 &      &      &       &      &      &       &       &      &      &       &      &      &       &       \\\hline
VSE$\infty$ \cite{chen2021learning}            & 79.7 & 96.4 & 98.9  & 64.8 & 91.4 & 96.3  & 527.5 & 58.3 & 85.3 & 92.3  & 42.4 & 72.7 & 83.2  & 434.3 \\
MMCA \cite{wei2020multi}           & 74.8 & 95.6 & 97.7  & 61.6 & 89.8 & 95.2  & 514.7 & 54.0 & 82.5 & 90.7  & 38.7 & 69.7 & 80.8  & 416.4 \\
CHAN\cite{pan2023fine}             & 81.4 & 96.9 & 98.9  & 63.8 & 90.4 & 95.8  & 525.0 & 59.8 & 87.2 & 93.3  & 44.9 & 74.5 & 84.2  & 443.9 \\
CORA\cite{pham2024composing}             & 82.8 & 97.3 & 99.0  & 67.3 & 92.4 & 96.9  & 535.6 & 64.3 & 87.5 & 93.6  & 45.4 & 74.7 & 84.6  & 450.1 \\
LAPS\cite{fu2024linguistic}             & 84.1 & 97.4 & 99.2 & 72.1 & 93.9 & 97.4 & 544.1 & 67.1 & 88.6 & 94.3 & 53.0 & 79.5 & 87.6 & 470.1 \\
HREM\cite{fu2023learning}             & 82.9 & 96.9 & 99.0  & 66.1 & 91.6 & 96.5  & 530.7 & 64.0 & 88.5 & 93.7  & 45.4 & 75.1 & 84.3  & 450.9 \\

TVRN[2024TMM]       & 81.1 & 96.4 & 98.8   & 67.7 & 92.3 & 97.1  & 533.4  & 61.1 & 86.3 & 92.5 & 45.0 & 75.0 & 84.8 & 445.2    \\

OS-HGAdapter(ours)           & 94.4 & 99.6 & 100.0 & 91.2 & 99.8 & 99.9  & \textcolor{red}{584.9} & 93.3 & 99.8 & 100.0 & 89.0 & 99.7 & 100.0 & \textcolor{red}{581.8} \\\hline
\end{tabular}
\label{table1}
}
\end{table*}

\section{Experiments}
\subsection{Datasets Descriptions}Our experiments utilize two benchmark datasets: Flickr30K \cite{young2014image} and MS-COCO \cite{chen2015microsoft}. The MS-COCO dataset comprises 123,287 images, which all annotated with five textual descriptions. Following the partitioning strategy established in \cite{huang2017instance,van2009visual,wei2020multi}, the dataset is divided into 113,287 training samples, 5,000 validation samples, and 5,000 test samples. For evaluation consistency, we report averaged metrics across five independent trials on a 1K subset of test images and additionally evaluate on 5K test set. The Flickr30K dataset contains 31,783 images sourced, each paired with five descriptive captions. Adhering to the protocol defined in \cite{huang2017instance}, we allocate 1,014 valid images, 1,000 for test sets, and retain the remaining samples for training purposes.
\subsection{Evaluation Metrics}
Our quantitative assessment framework adopts top-K retrieval precision as the core evaluative criterion, rigorously matching the success ratio of query-to-candidate alignments within the closest K-level retrieval candidates. Higher metric values directly indicate better model performance. To fully describe the model's matching ability, we divide the top 10 retrieval results into three different levels and summarize the mutual retrieval performance of image-to-text and text-to-image modes to provide the performance summary.

\begin{table}[h]
\setlength{\tabcolsep}{1mm}
\centering
\small
\caption{Test Results of different methods on Flickr30K test set. Use \textcolor{red}{red} to highlight the best RSUM.}
\begin{tabular}{@{}llllllll@{}}\hline
\multirow{3}{*}{Methods} & \multicolumn{7}{c}{Flickr30K test set}           \\\hline
 & \multicolumn{3}{l}{I-\textgreater{}T} & \multicolumn{3}{l}{T-\textgreater{}I} & \multirow{2}{*}{RSUM} \\
                        & R@1  & R@5  & R@10 & R@1  & R@5  & R@10  &       \\\hline
\textbf{Region+BiGRU}            &      &      &      &      &      &       &       \\\hline
VSE$\infty$\cite{chen2021learning}       & 77.1 & 94.5 & 97.1 & 58.5 & 84.1 & 89.6  & 500.9 \\
SCAN\cite{li2019visual}       & 67.4 & 90.3 & 95.8 & 48.6 & 77.7 & 85.2  & 465.0 \\
VSRN\cite{li2019visual} & 71.3 & 90.6 & 96.0 & 54.7 & 81.8 & 88.2  & 482.6 \\
HREM\cite{fu2023learning}      & 81.4 & 96.5 & 98.5 & 60.9 & 85.6 & 91.3  & 514.3 \\
CHAN\cite{pan2023fine}      & 79.7 & 94.5 & 97.3 & 60.2 & 85.3 & 90.7  & 507.8 \\
ESSE \cite{Wang2024Estimating}         & 80.2 & 94.6 & 97.2   & 60.9 & 85.6 & 90.87  & 509.3 \\  

CORA\cite{pham2024composing}            & 82.3 & 96.1 & 98.0   & 63.0 & 87.4 & 92.8  & 519.6 \\
 
CSRC\cite{Li2023Commonsense}       & 79.6 & 96.2 & 99.1   & 67.3 & 91.3 & 96.7  & 530.2 \\

OS-HGAdapter       & 93.1 & 98.8 & 99.9 & 84.1 & 97.6 & 100.0 & \textcolor{red}{573.5} \\ \hline
\textbf{Region+BERT}             &      &      &      &      &      &       &       \\\hline
VSE$\infty$\cite{chen2021learning}        & 81.7 & 95.4 & 97.6 & 61.4 & 85.9 & 91.5  & 513.5 \\
VSRN\cite{li2019visual}     & 79.2 & 94.6 & 97.5 & 60.6 & 85.6 & 91.4  & 508.9 \\
HREM\cite{fu2023learning}         & 84.0 & 96.1 & 98.6 & 64.4 & 88.0 & 93.1  & 524.2 \\
CHAN\cite{pan2023fine}         & 80.6 & 96.1 & 97.8 & 63.9 & 87.5 & 92.6  & 518.5 \\
CORA\cite{pham2024composing}              & 83.4 & 95.9 & 98.6  & 64.1 & 88.1 & 93.1  & 523.3 \\
LAPS\cite{fu2024linguistic}         & 85.1  & 97.7 & 99.2 & 74.0 & 93.0 & 96.3 & 545.3 \\
 
FF \cite{Wu2023Feature}         & 86.2 & 97.3 & 99.1   & 67.7 & 90.2 & 94.6  & 535.1 \\   
OS-HGAdapter      & 94.6 & 99.8 & 100  & 90.6 & 99.8 & 99.9  & \textcolor{red}{584.7} \\ \hline

\end{tabular}
\label{table2}
\end{table}

\subsection{Implementation Details} 
\begin{figure}[h]
\centering
\small
\includegraphics[width=1\columnwidth]{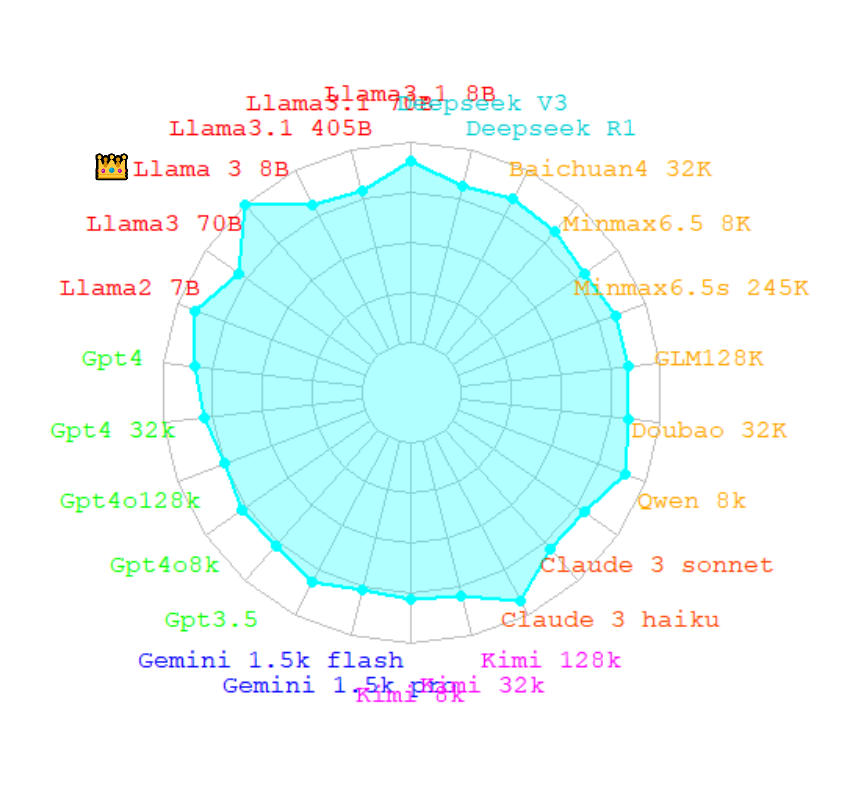} 
\caption{The average information entropy of various large language models (e.g. GPT4, Claude 3 and deepseek R1) after prompt learning and entropy increase on the COCO dataset}
\label{Figure6}
\end{figure}

We utilize Llama-3-8B-Instruct, fine-tuned based on community feedback, to extract open semantic entropy. This enhances the stability and responsiveness of the response, effectively supporting the required prompt word library. As shown in Fig. \ref{Figure6}, Llama-3-8B demonstrates the highest average information content for designed prompts to increase open semantic entropy in the COCO dataset, outperforming high-parameter models and its evolved version Llama-3.1. Although this does not imply superior overall performance, Llama-3-8B excels in expanding semantic space and information volume, better fitting open-world data. During training, we used two NVIDIA GeForce RTX 4090 GPUs, one for Llama-3-8B data enhancement and the other for training the cross-modal alignment network. We provide semantic enforced entropy data of COCO and flickor30k caption at: \href{https://github.com/multimodel-learner/OV-HGAdapter-dataset/}{https://github.com/multimodel-learner/OV-HGAdapter-dataset/
}.

\subsubsection{Quantitative comparison}In our study, OS-HGAdapter was benchmarked against leading methods on Flickr30K and COCO datasets. Aligning with CHAN's evaluation strategy, we present results from a single model without ensemble techniques. Table \ref{table1} highlights OS-HGAdapter's significant performance improvements on the COCO 5K and 5-fold 1K datasets. While most methods struggle with the larger and more complex MS-COCO 5K set, OS-HGAdapter maintains accuracy. This decline in other methods is attributed to embedding space disorder caused by single-token encoding, worsened by data volume. OS-HGAdapter integrates token connections and leverages semantic entropy to enhance synonym understanding and encoding order. Unlike conventional methods that conflate irrelevant information, OS-HGAdapter ensures R@5 retrieval accuracy closely aligns with actual values.

As shown in Table \ref{table2}, OS-HGAdapter outperforms previous methods, with BiGRU-based and BERT-based configurations achieving RSUMs of 573.5 and 592.5, respectively. The BiGRU-based model surpasses the CHAN baseline, improving bidirectional R@1 retrieval by over 16.8\% and 40.1\%. The BERT-based variant excels by 14\% in RSUM, demonstrating superior performance. Additionally, the performance gap between text-image and image-text retrieval has been significantly reduced. Post-entropy optimization, the BiGRU-based model narrows the gap from 32\% to 10.7\%, while the BERT-based model reduces it to just 2\%, validating the effectiveness of modal entropy in enhancing retrieval outcomes.

\begin{figure*}[t]
\centering
\small
\includegraphics[width=15cm]{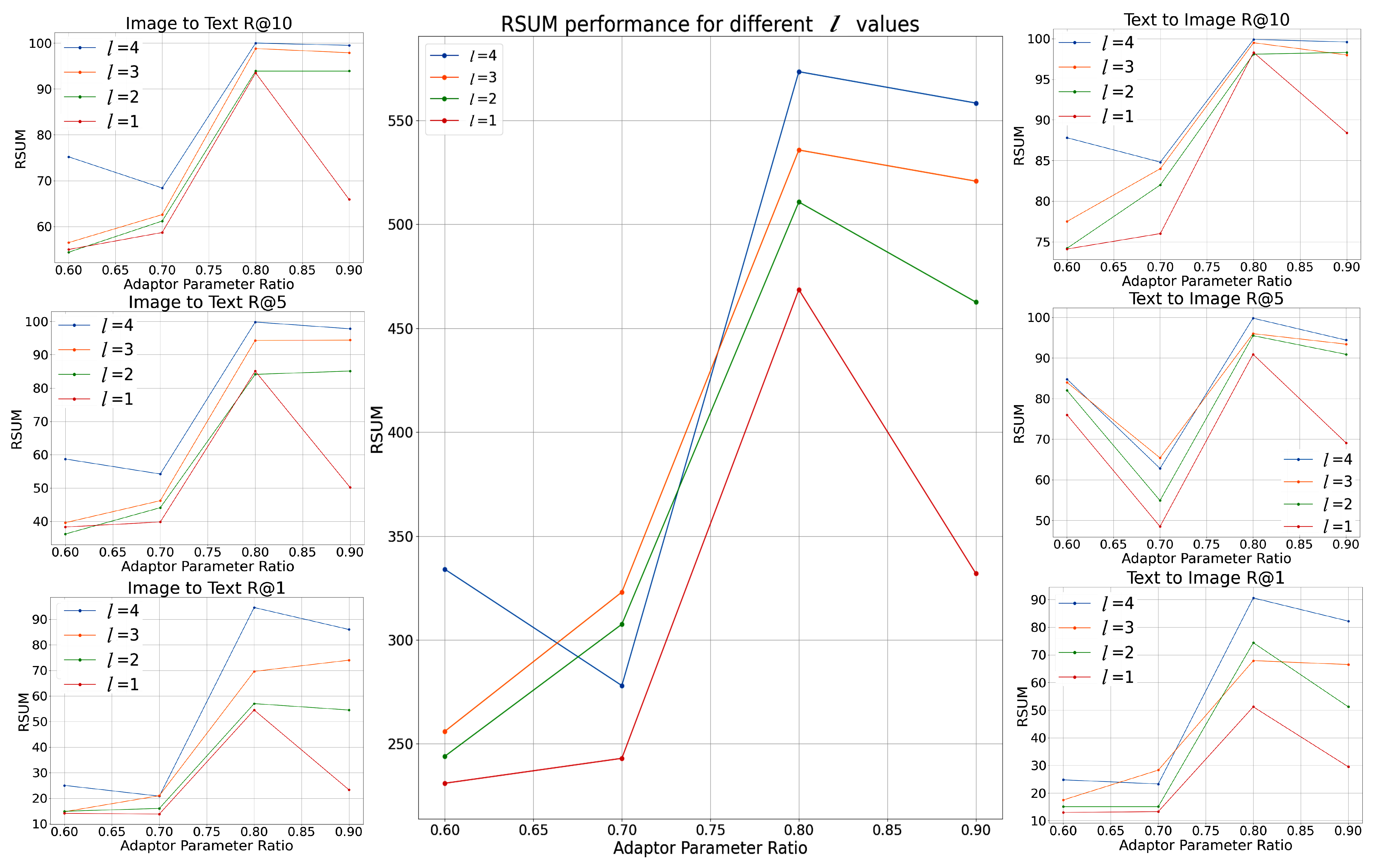} 
\caption{Model performance indicators RSUM and R@1, R@5, R@10 on the COCO dataset under different $l$ and different hypergraph adapter ratios $\alpha$ .}
\label{Figure7}
\end{figure*}

Table \ref{table1} highlights OS-HGAdapter's significant performance improvements on the COCO 5K and 5-fold 1K datasets. While most methods struggle with the larger and more complex MS-COCO 5K set, OS-HGAdapter maintains accuracy. This decline in other methods is attributed to embedding space disorder caused by single-token encoding, worsened by data volume. OS-HGAdapter integrates token connections and leverages semantic entropy to enhance synonym understanding and encoding order. Unlike conventional methods that conflate irrelevant information, OS-HGAdapter ensures R@5 retrieval accuracy closely aligns with actual values.

\begin{figure}[!ht]
  \centering
  \subfloat[Visual adapter with different $\beta$ values on RSUM]
  {\includegraphics[width=0.5\linewidth]{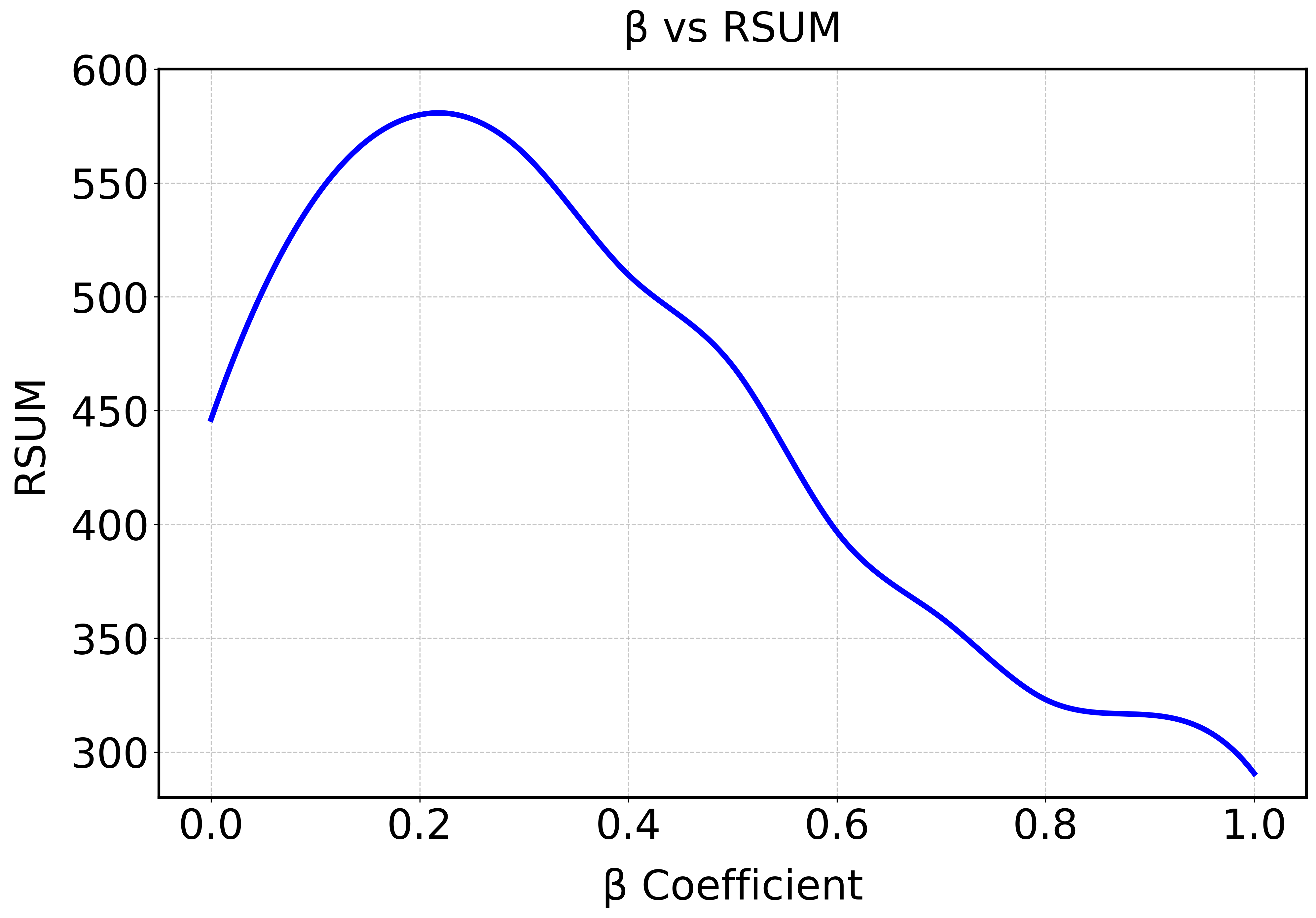}\label{fig:subfig1}}
  \subfloat[Visual adapter with different $\beta$ values on gredienct deviation]
  {\includegraphics[width=0.5\linewidth]{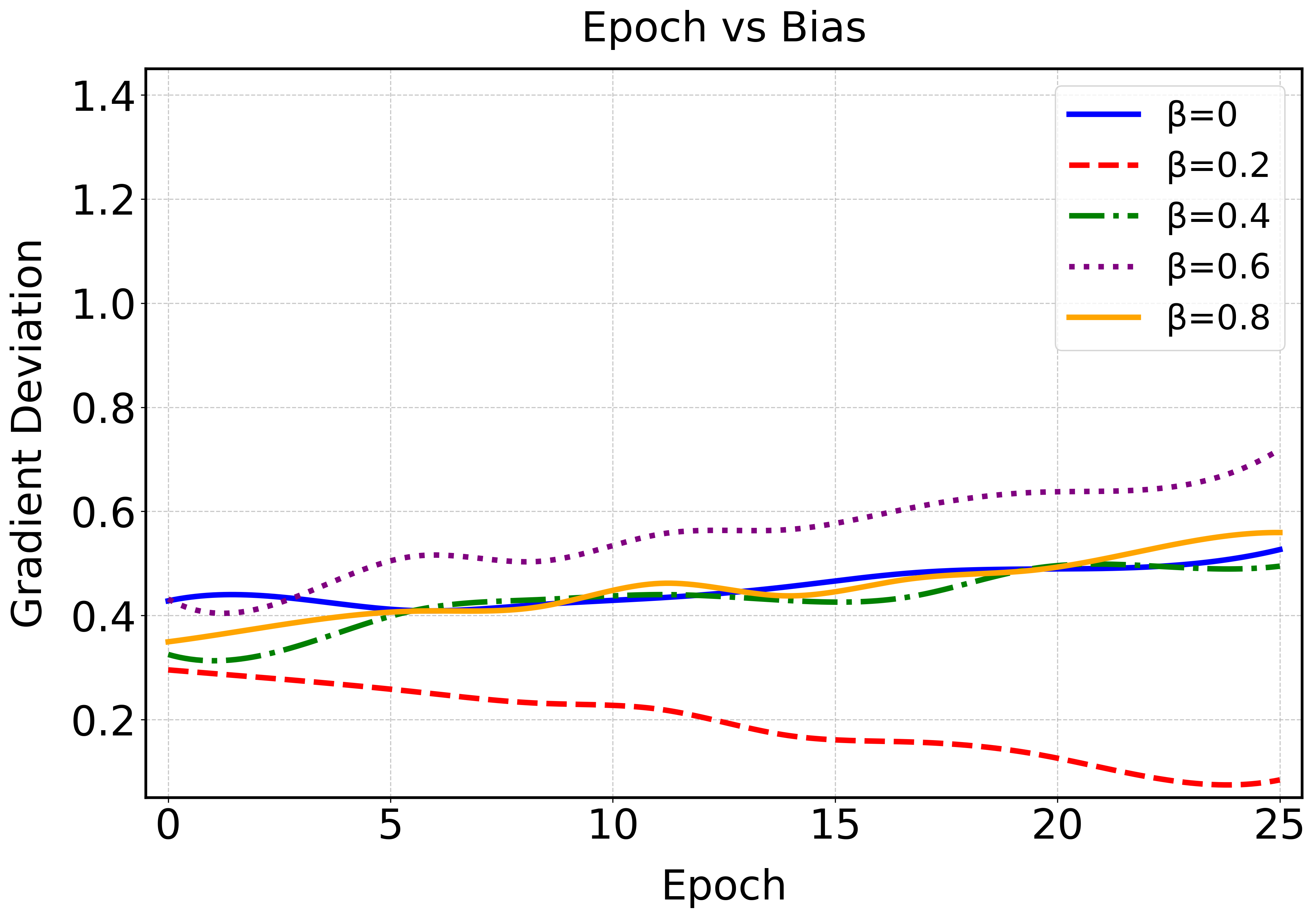}\label{fig:subfig2}}
\end{figure}

\begin{figure*}[htbp]
    \centering
    \begin{minipage}[b]{1\textwidth} 
        \includegraphics[width=14cm,trim= 0 350 0 0 ,clip]{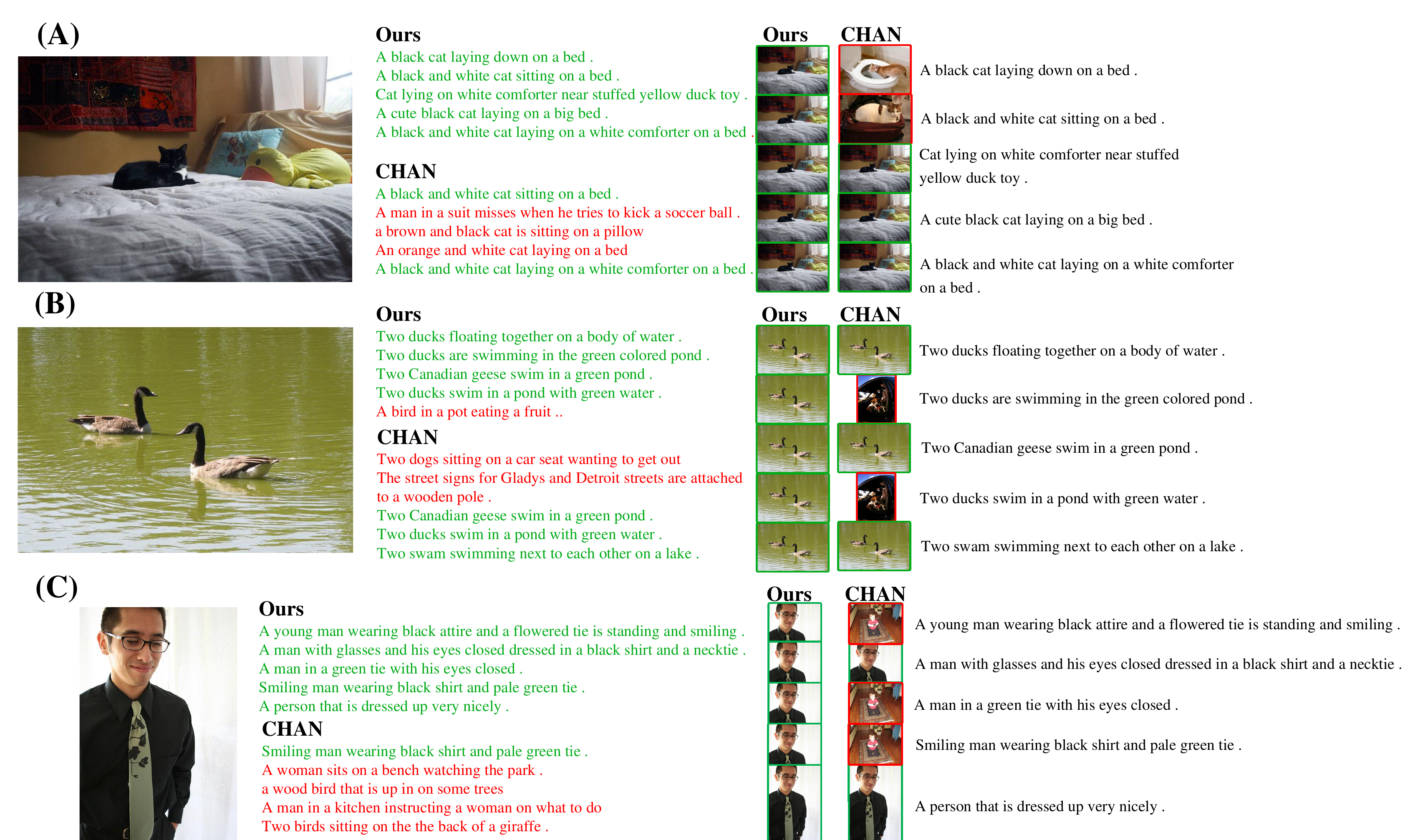}
        \centering
        \caption{The mutual search results of two related pictures and texts compare with the baseline CHAN\cite{pan2023fine}. Green fonts are used to indicate that the picture-to-text search is ground-truth, and red fonts indicate search errors. Green boxes indicate that the text-to-picture search is ground-truth, and red boxes indicate search errors.}
        \label{Figure8}
    \end{minipage}
    \hfill 
    \begin{minipage}[b]{1\textwidth}
        \includegraphics[width=0.7\textwidth]{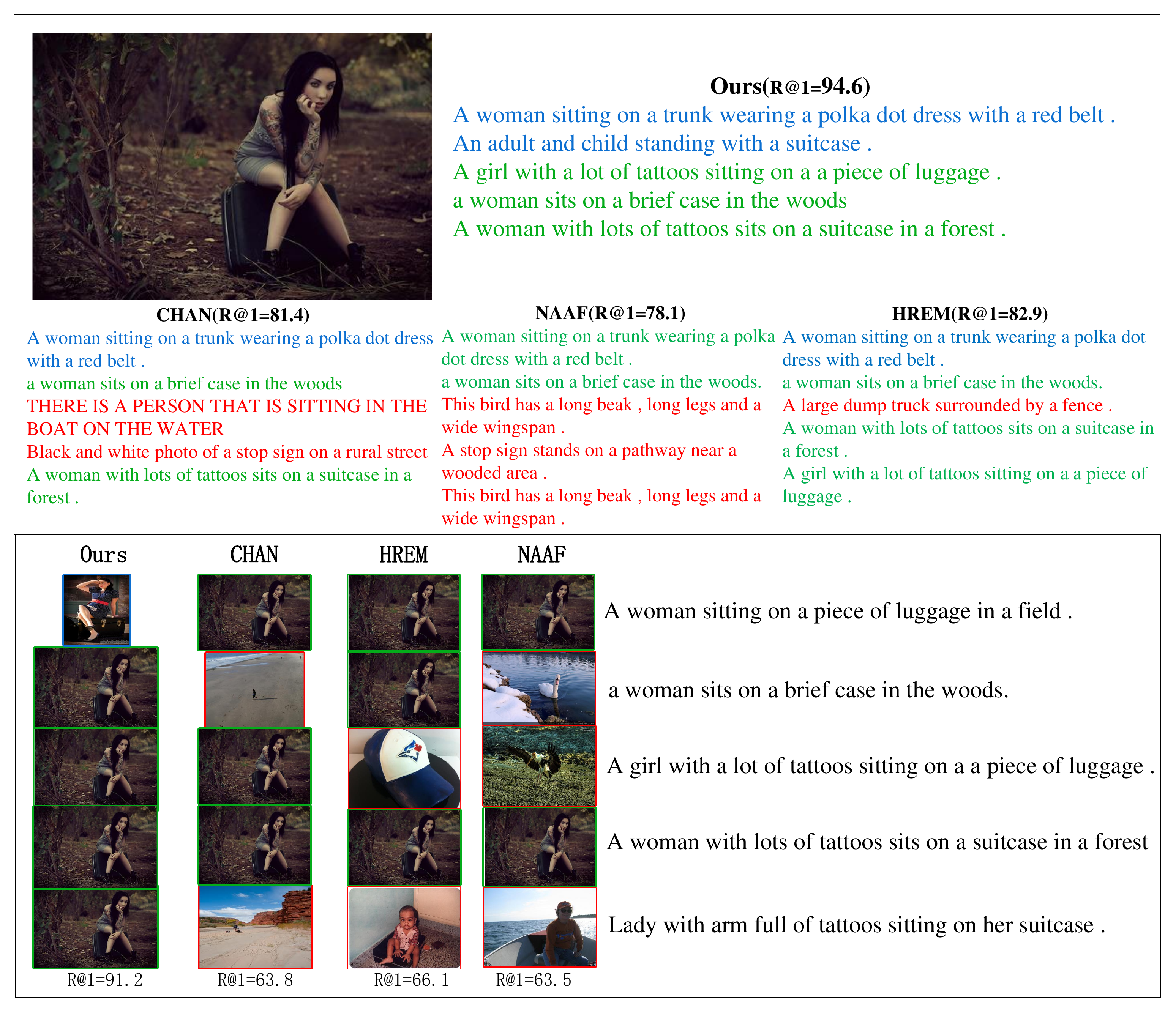}
        \centering
        \caption{The results of image-text mutual retrieval when the text embedding space is disordered with three models CHAN\cite{pan2023fine}, NAAF\cite{Zhang2022CVPR}, HREM\cite{fu2023learning}. Red text shows retrieval errors, and blue text represents semantically similar sentences to the ground truth. Green boxes signify accurate text-to-image matches, red boxes highlight retrieval errors, and blue boxes indicate images with similar semantics to the ground truth.}
        \label{Figure9}
    \end{minipage}
\end{figure*}

\section{Further Analysis}

\subsection{Hyperparameter analysis} We varied the number of synonymous sentences $l$ to control open semantic entropy and adjusted the adapter ratio to study the impact of hypergraph information on calibration. As shown in Fig. \ref{Figure7}, the optimal performance was achieved with $\alpha$ = 0.2, which count by the NMI-based fusion ratio and $l$ = 4. Notably, the size of open information entropy and the hypergraph ratio must be balanced: excessive open connections can distort the embedding space, leading to overfitting or a retrieval rate of zero.

\subsection{Ablation Study}
To architecturally dissect OS-HGAdapter's component efficacy, we execute systematic ablation experiments on the Flickr30K benchmark dataset under default parameter initialization, with the BiGRU-architected CHAN serving as the control variate for comparative analysis.

\begin{table}[!ht]
\setlength{\tabcolsep}{0.5mm}
\scriptsize
\caption{Ablation studies on Flickr30K test set with the BiGRU-base CHAN as the baseline}

\begin{tabular}{@{}lllllllll@{}}
\multirow{3}{*}{Adapter type} & \multicolumn{3}{c}{IMG$\rightarrow$TEXT} & \multicolumn{3}{c}{TEXT$\rightarrow$IMG} & \multirow{2}{*}{RSUM} & \multirow{2}{*}{Params} \\ 

& R@1  & R@5  & R@10 & R@1  & R@5  & R@10  &        &       \\\hline
CHAN\cite{pan2023fine}(baseline)     & 79.7 & 94.5 & 97.3 & 60.2 & 85.3 & 90.7  & 507.8 & 112,367,872 \\        
+Avgadapter\cite{Zeiler2013ICLR} & 75.2 & 96.7 & 99.2 & 67.2 & 92.5 & 97.0  & 527.9 & 112,367,872          \\
+Maxadapter\cite{Tolias2016Particular} & 76.2 & 94.0 & 97.4 & 62.6 & 84.0 & 90.9  & 505.6 & 112,367,872          \\
+GCNadapter\cite{Kipf2016} & 47.9  & 81.8 & 92.2 & 40.7 & 76.2 & 87.1 & 425.84 & 112,385,288 \\
+\textbf{OS-HGAdapter (ours)} & 93.1 & 98.8 & 99.9 & 84.1 & 97.6 & 100.0 & \textcolor{red}{573.5} & 112,395,997 \\\hline
\end{tabular}

\label{table3}
\end{table}

The hypergraph adapter structure is the core component of our experiment. In Table \ref{table3}, we designed various feature adapters to validate its effectiveness. Through testing different adapter kernels, we found that the Maxpooling kernel approaches but does not surpass CHAN's performance. In contrast, the Avgpooling kernel achieves a 3.9\% improvement over the baseline, demonstrating the benefits of open information entropy. We also designed GCNadapter based on the structure of \cite{Kipf2016}, However, because the GCN map construction process does not have the special concat structure of HGNN, its effect is not as good as other adapters. Our hypergraph adapter, leveraging semantic multilateral connections and increased open semantic information entropy, significantly outperforms other methods, proving the efficacy of the proposed network structure.

\subsection{Effect of the network architecture}

\begin{table}[!ht]
    \setlength{\tabcolsep}{0.1mm}
    \centering
    \small
    \caption{The R@1 retrieval results for different adapter configurations on the COCO and Flickr30k datasets. × and \cmark respectively indicate whether to use the corresponding adapter. Rank@1 indicates the accuracy rate of the search ranked first.}
    \begin{tabular}{@{}lllll@{}}
    \hline
         \multirow{2}{*}{Dataset} & \multirow{2}{*}{Text Adapter} & \multirow{2}{*}{Vision Adapter} & Image-to-Text & Text-to-Image \\ 
       ~ & ~ & ~ & Rank@1 & Rank@1 \\ \hline
        \multirow{3}{*}{COCO} & \cmark & × & 3.7 & 52.7 \\ 
        ~ & × & \cmark & 53.9 & 5.6 \\ 
        ~ & \cmark & \cmark & \textcolor{red}{94.4} & \textcolor{red}{91.2} \\ \hline
        \multirow{3}{*}{Flickor} & \cmark & × & 4.8 & 54.8 \\ 
        ~ & × & \cmark & 53.6 & 4.4 \\ 
        ~ & \cmark & \cmark & \textcolor{red}{94.6} & \textcolor{red}{90.6} \\ \hline
    \end{tabular}
    \label{table4}
\end{table}

Table \ref{table4} compares the impact of visual and textual modalities on the alignment results. When using a single visual adapter or single text adapter, it can only make the single modal retrieval still effective. Only when a bimodal adapter is utilized can the embedding space be calibrated in the correct direction and thus the significantly improved results.

We also use different $\beta$ values to observe RSUM and gradient deviation during training. We use formula \ref{dev} to calculate the gradient deviation. The experiment proves that when the entropy value of the visual adapter converges with that of the text adapter, that is, when the ratio of the original data to the hypergraph processed data is the same, the RSUM of the model increases, and the gradient can be directed to the normal value during training.

\subsection{Case Study}
To empirically validate the architectural superiority of OS-HGAdapter on the COCO dataset, we analyzed its retrieval results. Leveraging entropy-enhancing fusion, our model accurately captures subtle differences in tokens and descriptions, avoiding embedding space mismatches. In Fig. \ref{Figure8}(A), although the sentence "playing football" is unrelated to "black cat," our synonym embedding strategy correctly distinguishes between white-orange and black-brown cats. In text-image retrieval, while two orange cats appear, our model correctly emphasizes the black cat, validating its accuracy.

In Fig. \ref{Figure8}(B), other methods incorrectly include an image of two dogs in a car in the retrieval results for two ducks, highlighting encoding overlap in the embedding space. In contrast, our model avoids such errors by effectively correcting encoding confusion.

In the incorrect example shown in Fig. \ref{Figure9}, methods without LLM and hypergraph adapter correction produce R@5 retrieval results with sentences unrelated to the image content, lacking keyword connections. In contrast, OS-HGAdapter maintains content consistency in image-to-text retrieval, accurately retrieving synonyms (e.g., "suitcase") even for less relevant results. Importantly, in text-to-image retrieval, our method retrieves more relevant matches and effectively calibrates the embedding space, even for imperfect results. By increasing entropy in the large language model, we mitigate dataset sparsity, improving both matching accuracy and embedding space precision.

\section{Conclusion}
This work introduces an innovative cross-modal learning framework which aims to alleviate the synonyms semantic gap between visual and textual representations while improving inter-modal alignment's accuracy and computational efficiency. We are the first to design and use a hypergraph adapter to efficiently deepen the understanding and encoding of multilateral semantic relations, achieving a deep understanding of synonymous sentences and efficient cross-modal alignment. Our innovative solution improves the accuracy and efficiency of image-text semantic consistency retrieval. Evaluations on dual datasets and ablation studies substantiate our model's efficacy. Future research will explore the interplay between entropy-enhanced retrieval and LLMs, as well as investigate the generalizability of artificially synthesized data across downstream tasks.


\begin{thebibliography}{00}

\bibitem{quiroga2005invariant}
R. Q. Quiroga, L. Reddy, G. Kreiman, C. Koch, and I. Fried, ``Invariant visual representation by single neurons in the human brain,'' Nature, vol. 435, no. 7045, pp. 1102--1107, 2005.

\bibitem{wei2022learning}
Y. Wei, D. Hu, Y. Tian, and X. Li, ``Learning in audio-visual context: A review, analysis, and new perspective,'' arXiv preprint arXiv:2208.09579, 2022.



\bibitem{wu2022characterizing}
N. Wu, S. Jastrzebski, K. Cho, and K. J. Geras, ``Characterizing and overcoming the greedy nature of learning in multi-modal deep neural networks,'' in International Conference on Machine Learning, 2022, pp. 24043--24055.

\bibitem{huang2022modality}
Y. Huang, J. Lin, C. Zhou, H. Yang, and L. Huang, ``Modality competition: What makes joint training of multi-modal network fail in deep learning?(provably),'' in International conference on machine learning, 2022, pp. 9226--9259.

\bibitem{fan2023pmr}
Y. Fan, W. Xu, H. Wang, J. Wang, and S. Guo, ``Pmr: Prototypical modal rebalance for multimodal learning,'' in Proceedings of the IEEE/CVF Conference on Computer Vision and Pattern Recognition, 2023, pp. 20029--20038.


\bibitem{frome2013devise}
A. Frome, G. S. Corrado, J. Shlens, S. Bengio, J. Dean, M. Ranzato, and T. Mikolov, ``Devise: A deep visual-semantic embedding model,'' in Advances in neural information processing systems, 2013, vol. 26.

\bibitem{lee2018stacked}
K. Lee, X. Chen, G. Hua, H. Hu, and X. He, ``Stacked cross attention for image-text matching,'' in Proceedings of the European conference on computer vision (ECCV), 2018, pp. 201--216.

\bibitem{li2022image}
K. Li, Y. Zhang, K. Li, Y. Li, and Y. Fu, ``Image-text embedding learning via visual and textual semantic reasoning,'' IEEE transactions on pattern analysis and machine intelligence, vol. 45, no. 1, pp. 641--656, 2022.

\bibitem{chen2021learning}
J. Chen, H. Hu, H. Wu, Y. Jiang, and C. Wang, ``Learning the best pooling strategy for visual semantic embedding,'' in Proceedings of the IEEE/CVF conference on computer vision and pattern recognition, 2021, pp. 15789--15798.

\bibitem{huang2017instance}
Y. Huang, W. Wang, and L. Wang, ``Instance-aware image and sentence matching with selective multimodal lstm,'' in Proceedings of the IEEE conference on computer vision and pattern recognition, 2017, pp. 2310--2318.

\bibitem{li2019visual}
K. Li, Y. Zhang, K. Li, Y. Li, and Y. Fu, ``Visual semantic reasoning for image-text matching'' in Proceedings of the IEEE/CVF international conference on computer vision, 2019, pp. 4654--4662.

\bibitem{Zhang2016Cross-modal}
Zhang L, Ma B, Li G, et al. Cross-modal retrieval using multiordered discriminative structured subspace learning[J]. IEEE Transactions on Multimedia, 2016, 19(6): 1220-1233.



\bibitem{diao2021similarity}
H. Diao, Y. Zhang, L. Ma, and H. Lu, ``Similarity reasoning and filtration for image-text matching,'' in Proceedings of the AAAI conference on artificial intelligence, 2021, vol. 35, no. 2, pp. 1218--1226.

\bibitem{liu2019focus}
C. Liu, Z. Mao, A. Liu, T. Zhang, B. Wang, and Y. Zhang, ``Focus your attention: A bidirectional focal attention network for image-text matching,'' in Proceedings of the 27th ACM international conference on multimedia, 2019, pp. 3--11.

\bibitem{wu2019learning}
Y. Wu, S. Wang, G. Song, and Q. Huang, ``Learning fragment self-attention embeddings for image-text matching,'' in Proceedings of the 27th ACM international conference on multimedia, 2019, pp. 2088--2096.

\bibitem{zhang2022negative}
K. Zhang, Z. Mao, Q. Wang, and Y. Zhang, ``Negative-aware attention framework for image-text matching,'' in Proceedings of the IEEE/CVF conference on computer vision and pattern recognition, 2022, pp. 15661--15670.




\bibitem{van2009visual}
J. C. Van Gemert, C. J. Veenman, A. W. M. Smeulders, and J. M. Geusebroek, ``Visual word ambiguity,'' IEEE transactions on pattern analysis and machine intelligence, vol. 32, no. 7, pp. 1271--1283, 2009.



\bibitem{chun2022eccv}
S. Chun, W. Kim, S. Park, M. Chang, and S. Oh, ``Eccv caption: Correcting false negatives by collecting machine-and-human-verified image-caption associations for ms-coco,'' in European Conference on Computer Vision, 2022, pp. 1--19.

\bibitem{parekh2020crisscrossed}
Z. Parekh, J. Baldridge, D. Cer, A. Waters, and Y. Yang, ``Crisscrossed captions: Extended intramodal and intermodal semantic similarity judgments for MS-COCO,'' arXiv preprint arXiv:2004.15020, 2020.

\bibitem{faghri2017vse++}
F. Faghri, D. J. Fleet, J. R. Kiros, and S. Fidler, ``Vse++: Improving visual-semantic embeddings with hard negatives,'' arXiv preprint arXiv:1707.05612, 2017.

\bibitem{gao2022hgnn+}
Y. Gao, Y. Feng, S. Ji, and R. Ji, ``HGNN+: General hypergraph neural networks,'' IEEE Transactions on Pattern Analysis and Machine Intelligence, vol. 45, no. 3, pp. 3181--3199, 2022.

\bibitem{feng2019hypergraph}
Y. Feng, H. You, Z. Zhang, R. Ji, and Y. Gao, ``Hypergraph neural networks,'' in Proceedings of the AAAI conference on artificial intelligence, 2019, vol. 33, no. 01, pp. 3558--3565.

\bibitem{lim2022hypergraph}
J. Lim, S. Yun, S. Park, and J. Y. Choi, ``Hypergraph-induced semantic tuplet loss for deep metric learning,'' in Proceedings of the IEEE/CVF Conference on Computer Vision and Pattern Recognition, 2022, pp. 212--222.



\bibitem{dong2022m5product}
X. Dong, X. Zhan, Y. Wu, Y. Wei, M. C. Kampffmeyer, X. Wei, M. Lu, Y. Wang, and X. Liang, ``M5product: Self-harmonized contrastive learning for e-commercial multi-modal pretraining,'' in Proceedings of the IEEE/CVF Conference on Computer Vision and Pattern Recognition, 2022, pp. 21252--21262.

\bibitem{young2014image}
P. Young, A. Lai, M. Hodosh, and J. Hockenmaier, ``From image descriptions to visual denotations: New similarity metrics for semantic inference over event descriptions,'' Transactions of the Association for Computational Linguistics, vol. 2, pp. 67--78, 2014.

\bibitem{chen2015microsoft}
X. Chen, H. Fang, T. Lin, R. Vedantam, S. Gupta, P. Doll{'a}r, and C. L. Zitnick, ``Microsoft coco captions: Data collection and evaluation server,'' arXiv preprint arXiv:1504.00325, 2015.

\bibitem{wei2020multi}
X. Wei, T. Zhang, Y. Li, Y. Zhang, and F. Wu, ``Multi-modality cross attention network for image and sentence matching,'' in Proceedings of the IEEE/CVF conference on computer vision and pattern recognition, 2020, pp. 10941--10950.

\bibitem{pan2023fine}
Z. Pan, F. Wu, and B. Zhang, ``Fine-grained image-text matching by cross-modal hard aligning network,'' in Proceedings of the IEEE/CVF conference on computer vision and pattern recognition, 2023, pp. 19275--19284.


\bibitem{Wang2024Estimating}
Wang Z, Gao Z, Han M, et al. Estimating the semantics via sector embedding for image-text retrieval[J]. IEEE Transactions on Multimedia, 2024.


\bibitem{cheng2022cross}
Y. Cheng, X. Zhu, J. Qian, F. Wen, and P. Liu, ``Cross-modal graph matching network for image-text retrieval,'' ACM Transactions on Multimedia Computing, Communications, and Applications (TOMM), vol. 18, no. 4, pp. 1--23, 2022.

\bibitem{chun2023improved}
S. Chun, ``Improved probabilistic image-text representations,'' arXiv preprint arXiv:2305.18171, 2023.

\bibitem{radford2021learning}
A. Radford, J. W. Kim, C. Hallacy, A. Ramesh, G. Goh, S. Agarwal, G. Sastry, A. Askell, P. Mishkin, J. Clark, and others, ``Learning transferable visual models from natural language supervision,'' in International conference on machine learning, 2021, pp. 8748--8763.

\bibitem{pratt2023does}
S. Pratt, I. Covert, R. Liu, and A. Farhadi, ``What does a platypus look like? generating customized prompts for zero-shot image classification,'' in Proceedings of the IEEE/CVF International Conference on Computer Vision, 2023, pp. 15691--15701.



\bibitem{cao2024madtp}
J. Cao, P. Ye, S. Li, C. Yu, Y. Tang, J. Lu, and T. Chen, ``MADTP: Multimodal Alignment-Guided Dynamic Token Pruning for Accelerating Vision-Language Transformer,'' in Proceedings of the IEEE/CVF Conference on Computer Vision and Pattern Recognition, 2024, pp. 15710--15719.

\bibitem{hu2021lora}
E. J. Hu, Y. Shen, P. Wallis, Z. Allen-Zhu, Y. Li, S. Wang, L. Wang, and W. Chen, ``Lora: Low-rank adaptation of large language models,'' arXiv preprint arXiv:2106.09685, 2021.

\bibitem{gao2024clip}
P. Gao, S. Geng, R. Zhang, T. Ma, R. Fang, Y. Zhang, H. Li, and Y. Qiao, ``Clip-adapter: Better vision-language models with feature adapters,'' International Journal of Computer Vision, vol. 132, no. 2, pp. 581--595, 2024.

\bibitem{wang2020cross}
S. Wang, R. Wang, W. Yao, S. Shan, and X. Chen, ``Cross-modal scene graph matching for relationship-aware image-text retrieval,'' in Proceedings of the IEEE/CVF winter conference on applications of computer vision, 2020, pp. 1508--1517.

\bibitem{wen2020learning}
K. Wen, X. Gu, and Q. Cheng, ``Learning dual semantic relations with graph attention for image-text matching,'' IEEE transactions on circuits and systems for video technology, vol. 31, no. 7, pp. 2866--2879, 2020.

\bibitem{chun2021probabilistic}
S. Chun, S. J. Oh, R. S. De Rezende, Y. Kalantidis, and D. Larlus, ``Probabilistic embeddings for cross-modal retrieval,'' in Proceedings of the IEEE/CVF Conference on Computer Vision and Pattern Recognition, 2021, pp. 8415--8424.

\bibitem{liu2021cross}
A. H. Liu, S. Jin, C. Lai, A. Rouditchenko, A. Oliva, and J. Glass, ``Cross-modal discrete representation learning,'' arXiv preprint arXiv:2106.05438, 2021.

\bibitem{vendrov2015order}
I. Vendrov, R. Kiros, S. Fidler, and R. Urtasun, ``Order-embeddings of images and language,'' arXiv preprint arXiv:1511.06361, 2015.

\bibitem{yang2022vision}
J. Yang, J. Duan, S. Tran, Y. Xu, S. Chanda, L. Chen, B. Zeng, T. Chilimbi, and J. Huang, ``Vision-language pre-training with triple contrastive learning,'' in Proceedings of the IEEE/CVF Conference on Computer Vision and Pattern Recognition, 2022, pp. 15671--15680.

\bibitem{bao2021beit}
H. Bao, L. Dong, S. Piao, and F. Wei, ``Beit: Bert pre-training of image transformers,'' arXiv preprint arXiv:2106.08254, 2021.

\bibitem{wang2022image}
W. Wang, H. Bao, L. Dong, J. Bjorck, Z. Peng, Q. Liu, K. Aggarwal, O. Mohammed, S. Singhal, S. Som, and others, ``Image as a foreign language: Beit pretraining for all vision and vision-language tasks,'' arXiv preprint arXiv:2208.10442, 2022.

\bibitem{petroni2019language}
F. Petroni, T. Rockt{"a}schel, P. Lewis, A. Bakhtin, Y. Wu, A. H. Miller, and S. Riedel, ``Language models as knowledge bases?'' arXiv preprint arXiv:1909.01066, 2019.

\bibitem{jiang2020can}
Z. Jiang, F. Xu, J. Araki, and G. Neubig, ``How can we know what language models know?'' Transactions of the Association for Computational Linguistics, vol. 8, pp. 423--438, 2020.

\bibitem{fu2023learning}
Z. Fu, Z. Mao, Y. Song, and Y. Zhang, ``Learning semantic relationship among instances for image-text matching,'' in Proceedings of the IEEE/CVF Conference on Computer Vision and Pattern Recognition, 2023, pp. 15159--15168.

\bibitem{shin2020autoprompt}
T. Shin, Y. Razeghi, R. L. Logan IV, E. Wallace, and S. Singh, ``Autoprompt: Eliciting knowledge from language models with automatically generated prompts,'' arXiv preprint arXiv:2010.15980, 2020.



\bibitem{zhou2022learning}
K. Zhou, J. Yang, C. C. Loy, and Z. Liu, ``Learning to prompt for vision-language models,'' International Journal of Computer Vision, vol. 130, no. 9, pp. 2337--2348, 2022.



\bibitem{chen2022plot}
G. Chen, W. Yao, X. Song, X. Li, Y. Rao, and K. Zhang, ``Plot: Prompt learning with optimal transport for vision-language models,'' arXiv preprint arXiv:2210.01253, 2022.

\bibitem{devlin2018bert}
J. Devlin, M. W. Chang, K. Lee, and K. Toutanova, ``Bert: Pre-training of deep bidirectional transformers for language understanding,'' arXiv preprint arXiv:1810.04805, 2018.

\bibitem{pham2024composing}
K. Pham, C. Huynh, S. Lim, and A. Shrivastava, ``Composing object relations and attributes for image-text matching,'' in Proceedings of the IEEE/CVF Conference on Computer Vision and Pattern Recognition, 2024, pp. 14354--14363.

\bibitem{fu2024linguistic}
Z. Fu, L. Zhang, H. Xia, and Z. Mao, ``Linguistic-Aware Patch Slimming Framework for Fine-grained Cross-Modal Alignment,'' in Proceedings of the IEEE/CVF Conference on Computer Vision and Pattern Recognition, 2024, pp. 26307--26316.



\bibitem{lim2022hypergraph}
J. Lim, S. Yun, S. Park, and J. Choi, ``Hypergraph-induced semantic tuplet loss for deep metric learning,'' in Proceedings of the IEEE/CVF Conference on Computer Vision and Pattern Recognition, 2022, pp. 212--222.

\bibitem{wang2023multilateral}
Z. Wang, Z. Gao, K. Guo, Y. Yang, X. Wang, and H. Shen, ``Multilateral semantic relations modeling for image text retrieval,'' in Proceedings of the IEEE/CVF Conference on Computer Vision and Pattern Recognition, 2023, pp. 2830--2839.


\bibitem{Li2023Commonsense}
Li W, Yang S, Li Q, et al. Commonsense-guided semantic and relational consistencies for image-text retrieval[J]. IEEE Transactions on Multimedia, 2023, 26: 1867-1880.

\bibitem{Ren2015Faster}
Shaoqing Ren, Kaiming He, Ross Girshick, and Jian Sun. Faster r-cnn: Towards real-time object detection with region proposal networks. NeurIPS, 28, 2015. 5

\bibitem{Krishna2017Visual}
Ranjay Krishna, Yuke Zhu, Oliver Groth, Justin Johnson, Kenji Hata, Joshua Kravitz, Stephanie Chen, Yannis Kalantidis, Li-Jia Li, David A Shamma, et al. Visual genome: Connecting language and vision using crowdsourced dense image annotations. IJCV, 123(1):32–73, 2017. 5

\bibitem{Anderson2018Bottom-up}
Peter Anderson, Xiaodong He, Chris Buehler, Damien Teney, Mark Johnson, Stephen Gould, and Lei Zhang. Bottom-up and top-down attention for image captioning and visual question answering. In CVPR, pages 6077–6086, 2018. 5, 7
\bibitem{Pang2024mutually}
Pang S, Zeng Y, Zhao J, et al. A mutually textual and visual refinement network for image-text matching[J]. IEEE Transactions on Multimedia, 2024.

\bibitem{Vaswani2017Attention}
Ashish Vaswani, Noam Shazeer, Niki Parmar, Jakob Uszkoreit, Llion Jones, Aidan N Gomez, Łukasz Kaiser, and Illia Polosukhin. Attention is all you need. NeurIPS, 30, 2017. 5

\bibitem{Vaswani2021Structured}
Xuri Ge, Fuhai Chen, Joemon M Jose, Zhilong Ji, Zhongqin Wu, and Xiao Liu. Structured multi-modal feature embedding and alignment for image-sentence retrieval. In ACMMM, pages 5185–5193, 2021. 5

\bibitem{Devlin2018Bert}
Jacob Devlin, Ming-Wei Chang, Kenton Lee, and Kristina Toutanova. Bert: Pre-training of deep bidirectional transformers for language understanding. arXiv preprint arXiv:1810.04805, 2018. 5, 7

\bibitem{Pennington2014Glove}
Jeffrey Pennington, Richard Socher, and Christopher D Manning. Glove: Global vectors for word representation. In EMNLP, pages 1532–1543, 2014. 5

\bibitem{Tolias2016Particular}
Tolias G, Sicre R, Jégou H. Particular Object Retrieval With Integral Max-Pooling of CNN Activations[C]//ICLR 2016-International Conference on Learning Representations. 2016: 1-12.



\bibitem{Zeiler2013ICLR}
Zeiler M D, Fergus R. Stochastic pooling for regularization of deep convolutional neural networks: 1st International Conference on Learning Representations, ICLR 2013[C]//1st International Conference on Learning Representations, ICLR 2013. 2013.

\bibitem{Kipf2016}
Kipf T N, Welling M. Semi-supervised classification with graph convolutional networks[J]. arXiv preprint arXiv:1609.02907, 2016.

\bibitem{Zhang2022CVPR}
Zhang K, Mao Z, Wang Q, et al. Negative-aware attention framework for image-text matching[C]//Proceedings of the IEEE/CVF conference on computer vision and pattern recognition. 2022: 15661-15670.

\bibitem{McDaid}
McDaid A F, Greene D, Hurley N, et al. Normalized Mutual Information to evaluate overlapping community finding algorithms[J].
\bibitem{Wu2023Feature}
Wu D, Li H, Gu C, et al. Feature first: Advancing image-text retrieval through improved visual features[J]. IEEE Transactions on Multimedia, 2023, 26: 3827-3841.


\end{thebibliography}
\end{document}